\documentclass{article}

\usepackage{microtype}
\usepackage{graphicx}
\usepackage{subcaption}
\usepackage{booktabs}

\usepackage{hyperref}
\usepackage{url}
\usepackage{algpseudocode}

\usepackage{graphicx}
\usepackage{multirow} 
\usepackage[table,xcdraw]{xcolor}
\usepackage{wrapfig}
\usepackage{arydshln}
\usepackage{booktabs} 
\usepackage{graphicx}
\usepackage[table]{xcolor}
\usepackage{caption}
\usepackage{bbding}   
\usepackage{pifont}   
\usepackage{amssymb}
\usepackage{float}
\usepackage{microtype}
\usepackage{pifont}

\makeatletter
\@namedef{ver@algorithmic.sty}{}
\makeatother

\usepackage[accepted]{icml2026}

\usepackage{amsmath}
\usepackage{amssymb}
\usepackage{mathtools}
\usepackage{amsthm}

\usepackage[capitalize,noabbrev]{cleveref}

\theoremstyle{plain}
\newtheorem{theorem}{Theorem}[section]
\newtheorem{proposition}[theorem]{Proposition}
\newtheorem{lemma}[theorem]{Lemma}
\newtheorem{corollary}[theorem]{Corollary}
\theoremstyle{definition}
\newtheorem{definition}[theorem]{Definition}

\theoremstyle{remark}
\newtheorem{remark}[theorem]{Remark}

\usepackage[textsize=tiny]{todonotes}

\icmltitlerunning{Beyond Logits: Coherent Hallucination Mitigation via Attention Contrastive Decoding}

\begin{document}

\twocolumn[
  \icmltitle{Beyond Logits: Coherent Hallucination Mitigation via \\ Attention Contrastive Decoding}

  \icmlsetsymbol{equal}{*}

  \begin{icmlauthorlist}
    \icmlauthor{Yujia Chen}{yyy}
    \icmlauthor{Rui Sun}{xxx}
    \icmlauthor{Huayu Mai}{yyy}
    \icmlauthor{Wangkai Li}{yyy}
            \icmlauthor{Zhangyu He}{yyy}
    \icmlauthor{Bingzhou Wang}{yyy}
    \icmlauthor{Aibing Li}{yyy}
    \icmlauthor{Wenzhang SUN}{comp}
    \icmlauthor{Tianzhu Zhang}{yyy}

  \end{icmlauthorlist}

  \icmlaffiliation{yyy}{School of Information Science and Technology / National Key Laboratory of Deep Space Exploration, University of Science and Technology of China, China}

  \icmlaffiliation{comp}{School of Software, Tsinghua University, China}
  \icmlaffiliation{xxx}{Shenzhen International Graduate School, Tsinghua University, China}

  \icmlcorrespondingauthor{Tianzhu Zhang}{tzzhang@ustc.edu.cn}

  \icmlkeywords{Machine Learning, ICML}

  \vskip 0.3in
]

\printAffiliationsAndNotice{}

\begin{abstract}

Large Vision-Language Models (LVLMs) demonstrate impressive multimodal 
capabilities, yet suffer from hallucination—generating factually inaccurate 
content. Contrastive Decoding (CD) mitigates this by contrasting amateur 
and expert branches at the logit level. However, our investigation reveals 
that such logit-level interventions fundamentally compromise generation 
coherence, necessitating restrictive penalty constraints unrelated to 
hallucination suppression.
We introduce Attention Contrastive Decoding (ACD), a training-free plug-in 
that complements logit-level CD by relocating part of the contrastive 
operations to the attention mechanism. Operating at an earlier stage of the 
forward pass, ACD performs smooth semantic-preserving interventions through 
an Adaptive Subtraction Strategy (ASS), which attenuates hallucination-associated 
attention patterns while amplifying critical visual information. Extensive 
experiments demonstrate that combining ACD with existing CD methods (\textit{e.g.,} 
VCD+ACD) produces substantially more coherent outputs with further reduced 
hallucinations, eliminating restrictive penalties while enabling trustworthy 
multimodal generation.Code is available in \href{https://github.com/jj-ccc/ACD}{https://github.com/jj-ccc/ACD}.

\end{abstract}

\section{Introduction}
\label{sec:intro}
Large Vision-Language Models (LVLMs)~\cite{liu2023improvedllava,liu2023llava,liu2024llavanext,zhu2023minigpt,Qwen-VL,li2023blip,driess2023palm,achiam2023gpt,chowdhery2023palm,alayrac2022flamingo} have emerged as a transformative paradigm in multimodal AI. By integrating visual encoders with large language models, LVLMs enable sophisticated cross-modal reasoning across diverse tasks, including image captioning~\cite{lin2014microsoft,li2023blip}, visual question answering~\cite{liu2023llava,liu2023improvedllava,liu2024llavanext,hudson2019gqa}, and complex multimodal reasoning~\cite{lu2022learn,alayrac2022flamingo,achiam2023gpt}. This advancement establishes a foundation for more natural and intuitive human-machine interaction.

\noindent Despite significant advances in LVLMs, the ``hallucination" phenomenon remains a critical challenge~\cite{li2023evaluating,fu2023challenger,yue2024less}. These hallucinations manifest as grammatically coherent outputs with factual inconsistencies relative to visual input. Unlike text-only LLM hallucinations, LVLM hallucinations exhibit distinct cross-modal characteristics, manifesting in three patterns: (1) \textit{Modal dissociation}: generation detached from visual input; (2) \textit{Perceptual fabrication}: hallucination of non-existent attributes (\textit{e.g.,} colors, quantities, spatial relations); (3) \textit{Semantic distortion}: inaccurate visual content summarization.
These cross-modal inconsistencies substantially undermine model credibility and pose significant safety risks in critical domains such as medical diagnostics and autonomous driving.

\begin{figure*}[t]  
\includegraphics[width=\textwidth]{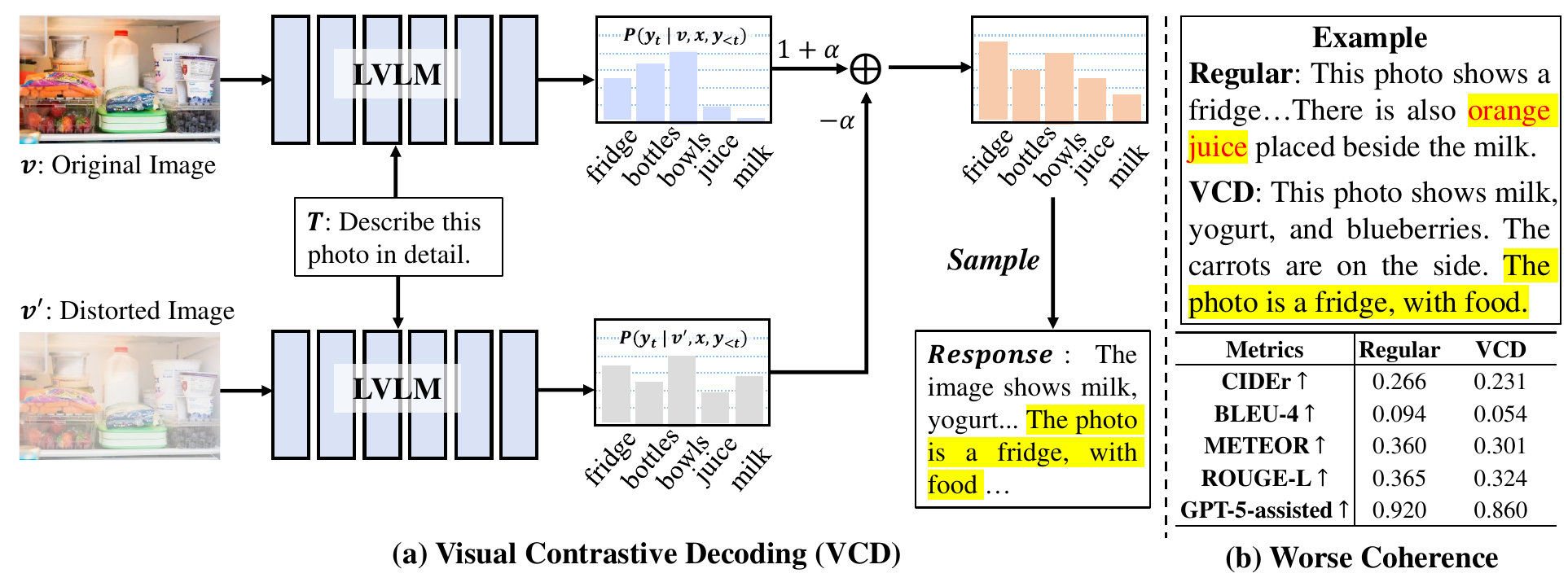} 

\caption{(a) VCD mitigates hallucinations by subtracting hallucination-induced 
distributions from original outputs. (b) However, logit-level subtraction 
disrupts linguistic coherence compared to vanilla decoding .}  
\label{fig1}  

\end{figure*}

\noindent Alleviating hallucinations in LVLMs has emerged as a critical research focus, with training-free Contrastive Decoding (CD) strategies~\cite{li2025mitigating,huo2024self,liu2024paying,an2025mitigating,leng2024mitigating,wang2024mitigating,chen2024halc} representing a prominent paradigm owing to their efficiency, flexibility, and strong interpretability. These approaches mitigate hallucinations through two principal mechanisms: hallucination induction and probability recalibration.
For hallucination induction, existing methods can be categorized into three primary classes: (1) Visual Contrastive Decoding (VCD)~\cite{leng2024mitigating} amplifies the model's reliance on linguistic priors by introducing Gaussian noise into visual inputs or completely eliminating visual signals~\cite{liu2024paying}, thereby constructing a hallucination-rich probability distribution. (2) Instructional Contrastive Decoding (ICD)~\cite{wang2024mitigating} employs adversarial prompts (\textit{e.g.,} ``you are a confused object detector'') to elicit hallucinated outputs. (3) Self-Introspective Decoding (SID)~\cite{huo2024self} strategically masks high-attention visual regions during decoding to provoke hallucination generation.
The probability recalibration mechanism then subtracts this artificially constructed hallucination distribution from the original generation probability. As illustrated in Figure~\ref{fig1}(a), this approach effectively suppresses factually inconsistent content while enhancing the model's fidelity to visual evidence, offering a convenient and theoretically interpretable solution for hallucination mitigation in LVLMs.

\noindent Despite substantial empirical evidence~\cite{leng2024mitigating,wang2024mitigating,huo2024self} demonstrating the efficacy of contrastive decoding (CD) strategies in mitigating LVLM hallucinations, our investigation reveals that CD approaches significantly compromise generation coherence—manifested as invalid token outputs, abrupt semantic transitions, and structural degradation—under specific operational conditions.
The fundamental limitation stems from CD's crude direct subtraction operation at the logits layer, which lacks nuanced regulation of the natural language generation process. Mathematically, this operation is equivalent to re-weighting the probability space via division, \textit{i.e.,} $P_{\text{final}} \propto P_{\text{original}} / P_{\text{distorted}}^\alpha$. While theoretically sound for suppressing hallucinations, such aggressive penalization induces a critical side effect known as the ``false positive'' problem~\cite{li2023contrastive}. Since the distorted model (negative view) retains its fundamental language modeling capabilities, it legitimately assigns high probabilities to high-frequency functional tokens (\textit{e.g.,} ``the'', ``of''). Consequently, direct logit subtraction indiscriminately penalizes these syntactically essential tokens, causing the decoding trajectory to deviate from the natural language manifold~\cite{liu2021dexperts}.

Furthermore, due to the inherent discontinuity and volatility of logits distributions, logit-level subtraction forcibly distorts vocabulary distribution probabilities, thereby disrupting the model's intrinsic language generation mechanisms. This direct intervention inadequately accounts for contextual dependencies and logical coherence, resulting in semantic fragmentation and structural collapse in generated outputs, as visualized in Figure~\ref{fig1}(b). Notably, since ICD and SID also operate through logit-level subtraction, they inevitably encounter identical limitations.
Our systematic experiments with LLaVA-1.5-7B~\cite{liu2023improvedllava} on the CHAIR benchmark~\cite{yue2024less} corroborate these observations: VCD consistently underperforms across multiple evaluation metrics, including CIDEr (description quality), METEOR (semantic comprehension), ROUGE-L (information completeness), and BLEU (generation precision). Additionally, GPT-5-assisted evaluation scores significantly trail those of conventional decoding methods. These empirical findings explicitly expose an inherent contradiction within current contrastive decoding paradigms—while logit-level contrastive operations effectively suppress hallucinations, they inevitably degrade linguistic quality, presenting a critical trade-off challenge for practical LVLM deployment.

\noindent Concurrently, probability recalibration mechanisms at the logits level frequently amplify tail token probabilities due to distribution discontinuities, resulting in invalid token generations. To mitigate this issue, existing contrastive decoding strategies typically incorporate adaptive penalty mechanisms to filter low-probability tokens. However, our in-depth analysis reveals that such compensatory mechanisms not only fail to adequately resolve the underlying issue but also introduce additional challenges.
\textbf{First, penalty mechanisms that mask low-probability tokens sacrifice decoding diversity} (constraining the model to high-probability tokens exclusively). This implies that penalty mechanisms function primarily to counteract CD's adverse effects rather than addressing the hallucination problem directly. This compensatory processing significantly restricts the richness and creativity of generated content, biasing outputs toward high-frequency, conventional expressions lacking natural linguistic variation and innovation. \textbf{Second, the intensity calibration of current penalty mechanisms predominantly relies on heuristic algorithms} (utilizing manually defined thresholds), severely limiting their adaptability and inevitably resulting in invalid token outputs. In diverse linguistic contexts and tasks, fixed penalty strategies inadequately accommodate variable generation requirements and cannot dynamically adjust filtering intensity based on contextual semantics, further exacerbating the inconsistency and instability of generated outputs.

\noindent To fundamentally address these limitations prevalent in existing CD methods, we propose Attention Contrastive Decoding (ACD), which relocates contrastive operations from the discrete logit space to the continuous attention layer.
This architectural choice leverages the attention mechanism's inherent capacity to weight input information and adapt to contextual cues, facilitating smooth information flow~\cite{vaswani2017attention,castin2023smooth}. Consequently, ACD naturally generates more coherent outputs while circumventing the need for penalty terms and preserving generation diversity (Detailed theoretical and empirical analyses are provided in the Appendix~\ref{sec:theory} and \ref{appendix:smoothness}). This adaptive contrastive mechanism at the attention level not only precisely localizes and suppresses hallucination sources but also preserves the model's intrinsic language generation capabilities, offering a more elegant and effective solution for hallucination mitigation in LVLMs.
Beyond architectural relocation, ACD introduces an Adaptive Subtraction 
Strategy (ASS) for finer-grained hallucination control. Unlike uniform 
subtraction in traditional CD, ASS distinguishes between two fundamentally 
different attention shift patterns under hallucination induction:(1)~\textbf{information degradation}, where decreased attention signals loss of focus on semantically relevant content that should be preserved, and (2)~\textbf{hallucination risk}, where increased attention indicates over-focus on spurious regions that warrant suppression.
Based on this assessment, ASS adaptively modulates the model's attention allocation across different regions. Specifically, it amplifies attention to areas exhibiting information degradation while suppressing regions with hallucination risk, thereby inhibiting hallucination generation at its source and enhancing focus on semantically valid content. This fine-grained control mechanism enables the model to maintain linguistic quality while improving effective information utilization.

\noindent In summary, our main contributions are three-fold:
(1) We reveal that logit-level subtraction in CD fundamentally compromises generation quality, and that penalty mechanisms fail to resolve invalid token generation while restricting output diversity.
(2) We propose Attention Contrastive Decoding (ACD), which relocates contrastive operations to the attention layer, leveraging attention's inherent continuity to preserve coherence without penalties. We further introduce the Adaptive Subtraction Strategy (ASS), which distinguishes information degradation from hallucination risk to achieve precise, selective suppression.
(3)Extensive experiments demonstrates ACD's capacity to generate more coherent, higher-quality content with significantly reduced hallucinations without penalty mechanisms, advancing the SOTA in reliable LVLMs decoding.

\begin{figure*}[!t]
\centering
\includegraphics[width=\textwidth]{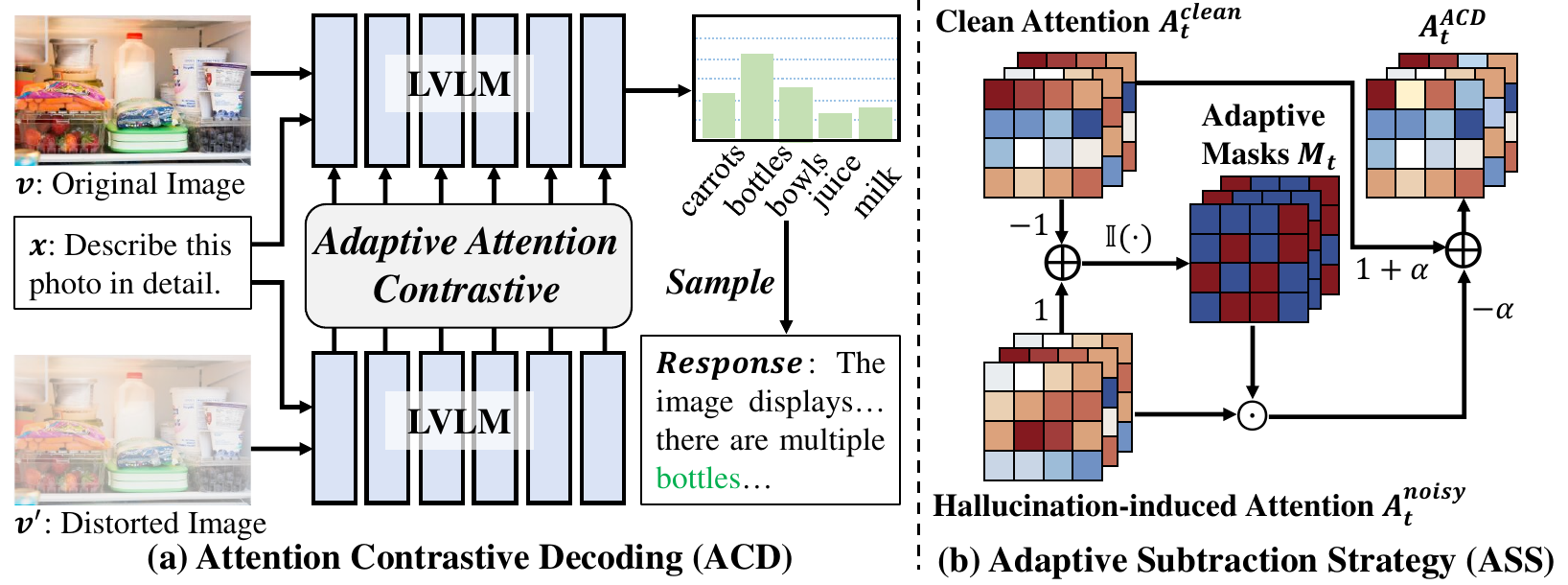}
\caption{\textbf{Overview of the proposed framework (VCD\,+\,ACD).}
\textbf{(a)~Attention Contrastive Decoding (ACD).}
Given an image $v$ and its perturbed counterpart $v'$, ACD computes 
parallel attention maps $A_t^{\text{clean}}$ and 
$A_t^{\text{noisy}}$ within the decoder. Contrastive subtraction is 
performed in the \textit{continuous attention space} 
(Equation~\ref{eq:ass}), producing a refined attention distribution 
$A_t^{\text{ACD}}$ that sharpens focus on visually grounded regions 
while suppressing hallucination-prone activations. The resulting 
logits are produced by the refined attention and decoded directly,
\textit{replacing} the logit-level subtraction and penalty mechanism
of conventional CD methods.
\textbf{(b)~Adaptive Subtraction Strategy (ASS).}
Rather than applying uniform subtraction, ASS constructs a binary 
mask $M_t$ that distinguishes \textit{signal-degraded} positions 
(amplified by $(1\!+\!\alpha)$) from \textit{hallucination-risk} 
positions (suppressed via contrastive subtraction), yielding a 
modulated attention map with an enhanced signal-to-noise ratio.}
 
\label{method}

\end{figure*}

\section{Related Work}

\subsection{Hallucination in LVLMs}

Research addressing hallucinations in large vision-language models (LVLMs) can be categorized into three primary approaches. 
\textit{First, contrastive decoding techniques} identify and suppress hallucinated content without parameter updates. Representative methods include VCD~\cite{leng2024mitigating}, ICD~\cite{wang2024mitigating}, and SID~\cite{huo2024self}, with advanced variants such as HALC~\cite{chen2024halc}, VaLiD~\cite{wang2024valid}, CMVED~\cite{li2025mitigating}, PAI~\cite{liu2024paying}, and AGLA~\cite{an2025mitigating} further refining these techniques through targeted attention manipulation.
\textit{Second, fine-tuning strategies} employ curated datasets incorporating negative~\cite{liu2023mitigating}, counterfactual~\cite{yu2024hallucidoctor}, or purified data~\cite{wang2024mitigating}. Related frameworks include HALVA~\cite{sarkar2024mitigating}, HACL~\cite{jiang2024hallucination}, PerturboLLaVA~\cite{chen2025perturbollava}, PATCH~\cite{shang2024pixels}, HIO~\cite{lyu2024alleviating}, Octopus~\cite{suo2025octopus}, OPA~\cite{yang2025mitigating}, TL-DPO~\cite{yoon2025stop}, and VASparse~\cite{zhuang2025vasparse}.
\textit{Third, direct intervention methods} manipulate visual features or attention mechanisms. These include DeCo~\cite{wang2024mllm}, VTI~\cite{liu2024reducing}, VDGD~\cite{ghosh2024visual}, ProjectAway~\cite{jiang2024interpreting}, TAME~\cite{tang2025intervening}, OPERA~\cite{huang2024opera}, VAR~\cite{jiang2025devils}, FarSight~\cite{tang2025seeing}, Nullu~\cite{yang2025nullu}, and ClearSight~\cite{yin2025clearsight}.
Notably, our method migrates contrastive decoding from logit level to attention layer, leveraging internal consistency for self-calibration rather than directly intervening in attention mechanisms.

\subsection{Contrastive Decoding in Hallucination}

Contrastive Decoding (CD) methods provide an elegant, training-free approach to mitigating hallucinations in LVLMs without parameter optimization. These strategies reduce hallucination generation by contrastively adjusting token probability distributions through two key mechanisms: hallucination induction and probability adjustment. Representative works include VCD~\cite{leng2024mitigating}, which injects Gaussian noise into visual inputs to construct hallucination-rich distributions; ICD~\cite{wang2024mitigating}, which employs negative prompts to induce hallucinations; and SID~\cite{huo2024self}, which strategically masks high-attention regions to expose model vulnerabilities. Advanced variants include HALC~\cite{chen2024halc} with automatic hallucination correction, VaLiD~\cite{wang2024valid} comparing early-layer outputs, CMVED~\cite{li2025mitigating} masking cross-modal attention, PAI~\cite{liu2024paying} removing visual components entirely, and AGLA~\cite{an2025mitigating} emphasizing critical regions through image-text matching models.
Several recent works also leverage attention mechanisms for hallucination mitigation, yet differ fundamentally from ACD. 
FarSight~\cite{tang2025seeing} and DAMRO~\cite{gong2024damro} modify attention within a single forward pass or use attention to select distorted tokens, but perform contrastive subtraction at the \textit{logit level}. 
HICD~\cite{jiang2025hicd} disperses attention to improve hallucination induction, again feeding results into logit-level subtraction. 
MoD~\cite{chen2025mixture} adaptively switches between addition and subtraction but operates entirely on logits. 
In contrast, ACD is the first to relocate the contrastive subtraction itself from the logit space to the attention layer, eliminating the coherence degradation and penalty mechanisms inherent in all logit-level approaches.
Despite demonstrated efficacy in hallucination mitigation, conventional CD strategies operate exclusively at the logits distribution level, substantially compromising output coherence and diminishing generative diversity.

\section{Method}
\label{sec:Method}

\subsection{Preliminary}

\textbf{Vanilla Decoding.}
A standard LVLM $\theta$ comprises a vision encoder, a 
vision-text interface, and an LLM decoder. Given an image $v$ 
and text query $x$, the model autoregressively generates 
output $y$ via:
\begin{small}
\begin{equation}
    y_t \sim p_{\theta}(y_t | v, x, y_{<t}) 
    \propto \exp\!\left(\operatorname{logit}_{\theta}
    (y_t | v, x, y_{<t})\right),
\end{equation}
\end{small}
where $y_t$ is the $t$-th generated token, $y_{<t}$ denotes 
all preceding tokens, and $\operatorname{logit}_{\theta}$ 
returns the pre-softmax scores. The attention weight matrix 
within each decoder head is:
\begin{small}
\begin{equation}
    A = \operatorname{Softmax}\!\left(
    \frac{Q K^\top}{\sqrt{d}}\right),
\end{equation}
\end{small}
where $Q, K \in \mathbb{R}^{L \times d}$ are the query and 
key matrices, $L$ is the sequence length, and $d$ is the 
head dimension. The output is then $AV$ with value matrix 
$V \in \mathbb{R}^{L \times d}$.

During autoregressive inference, the model caches $K$ and $V$ 
from all previous steps. The attention at step $t$ reduces to:
\begin{small}
\begin{equation}
    a_t = \operatorname{Softmax}\!\left(
    \frac{q_t K_{\leq t}^\top}{\sqrt{d}}\right),
\end{equation}
\end{small}
where $q_t \in \mathbb{R}^{1 \times d}$ is the current query 
and $K_{\leq t}$ aggregates cached keys. This cached attention 
computation serves as the intervention point for our proposed ACD.

\noindent \textbf{Contrastive Decoding.}
CD~\cite{leng2024mitigating,wang2024mitigating,huo2024self} 
mitigates hallucinations by subtracting a hallucination-rich distribution 
from the original output. The hallucination distribution is induced by 
perturbing inputs (visual corruption $v'$~\cite{leng2024mitigating}, 
adversarial prompts $x'$~\cite{wang2024mitigating}, or attention 
masking~\cite{huo2024self}).
Subsequently, a new contrastive probability distribution is computed by leveraging the differences between these two distributions. This contrastive distribution, denoted as $p_{cd}$, is defined as:
\begin{small}
\begin{equation}
\label{eq:3}
\begin{aligned}
p_{cd}\left(y \mid v, v', x'\right) = \operatorname{softmax} & \left[  (1+\alpha) 
\operatorname{logit}_\theta\left(y \mid v, x\right) \right. \\ 
& \left.-  \alpha \operatorname{logit}_\theta\left(y \mid v', x'\right)\right],
\end{aligned}
\end{equation}
\end{small}where larger $\alpha$ values indicate a stronger amplification of differences between the two distributions ($\alpha=0$ degenerates to Vanilla decoding).
A fundamental limitation of Equation~\ref{eq:3} is that the CD subtraction operation can drive certain reasonably token logits to extremely small or negative values, causing unreasonably low-probability tokens to emerge after softmax normalization and thus amplifying tail token probabilities. This necessitates a uniform penalization constraint that suppresses all low-probability outputs from the original inputs:
\begin{small}
\begin{equation}
\begin{aligned}
\label{pen}
& \mathcal{V}_{\text{head}}\left(y_{<t}\right) = \{y_t \in \mathcal{V}: \\ & p_{\theta}\left(y_t \mid v,x,y_{<t}\right)  \geq \beta \max _w p_{\theta}\left(w\mid v,x,y_{<t}\right)\}, \\
 & p_{cd}\left(y_t \mid v, v', x'\right)= 0, \textit{if}\, y_t \notin \mathcal{V}_{\text{head}}\left(y_{<t}\right),
\end{aligned}
\end{equation}
\end{small}where $\mathcal{V}$ is the output vocabulary of LVLMs and $\beta$ is a hyperparameter in $[0,1]$ for controlling the truncation of the next token distribution. Larger $\beta$ indicates more aggressive truncation, keeping only high-probability tokens. 
Combining the CD and the  plausibility constraint, we obtain the full formulation:
\begin{small}
\begin{equation}
\begin{aligned}
\label{eqcd}
 & y_t  \sim  \operatorname{softmax} \left[(1+\alpha) \operatorname{logit}_\theta\left(y_t \mid v, x, y_{<t}\right) \right. \\ & \left. - \alpha  \operatorname{logit}_\theta\left(y_t \mid v', x', y_{<t}\right)\right], 
\textit{subject} \, \textit{to} \, y_t \in \mathcal{V}_{\text{head}}\left(y_{<t}\right).
\end{aligned}
\end{equation}
\end{small}

\subsection{Attention Contrastive Decoding (ACD)}
\label{sec:acd}

Existing CD methods (Equation~\ref{eqcd}) perform contrastive 
subtraction in the discrete logit space, where even small perturbations 
can cause abrupt redistributions over the vocabulary, degrading 
linguistic coherence and necessitating ad-hoc penalty terms 
(Equation~\ref{pen}). We propose \textbf{Attention Contrastive 
Decoding (ACD)}, a lightweight plug-in that introduces contrastive 
operations into the \textit{continuous attention layer}, complementing 
logit-level CD to jointly suppress hallucinations. The complete 
framework is illustrated in Figure~\ref{method}.

\paragraph{Motivation.}
Attention weights reside in a continuous probability simplex produced 
by the Softmax operator, where arithmetic combinations yield 
semantically smooth transitions~\cite{vaswani2017attention,
castin2023smooth}. In contrast, logit subtraction operates on 
unbounded pre-softmax scores whose exponentiation amplifies minor 
differences into drastic probability shifts. By intervening before 
the logit stage, ACD modulates \textit{where} the model attends 
rather than \textit{what} it outputs, inherently preserving the 
linguistic structure of the downstream distribution.

\paragraph{Formulation.}
At decoding step $t$, we compute attention weights under two input 
configurations—the original input $(v, x)$ and a 
hallucination-inducing input $(v', x')$:
\begin{small}
\begin{equation}
\begin{aligned}
A_t^{\text{clean}} &= \mathrm{Softmax}\!\left( \frac{q_t \, {K_{\leq t}^{\text{clean}}}^{\!\top}}{\sqrt{d}} \right), \\
A_t^{\text{noisy}} &= \mathrm{Softmax}\!\left( \frac{q_t \, {K_{\leq t}^{\text{noisy}}}^{\!\top}}{\sqrt{d}} \right),
\end{aligned}
\end{equation}
\end{small}
where $q_t \!\in\! \mathbb{R}^{1 \times d}$ is the query at step 
$t$, and $K_{\leq t}^{\text{clean}},\, K_{\leq t}^{\text{noisy}} 
\!\in\! \mathbb{R}^{t \times d}$ are cached key matrices under the 
respective inputs. Unlike logit-level CD (Equation~\ref{eqcd}), ACD 
performs contrastive subtraction directly on attention weights:
\begin{small}
\begin{equation}
\label{eq:acd_basic}
A_t^{\text{base}} 
= (1 + \alpha)\, A_t^{\text{clean}} 
  - \alpha\, A_t^{\text{noisy}},
\end{equation}
\end{small}
where $\alpha > 0$ controls the contrastive strength. This operation 
amplifies attention on regions consistently highlighted by the clean 
model while suppressing those emphasized only under noise. For 
generality, we apply ACD uniformly across all decoder layers and 
attention heads.

\paragraph{Compatibility with Existing CD.}
ACD serves as a modular alternative to the logit-level 
subtraction and penalty mechanism shared by existing CD 
methods. Given any hallucination induction strategy 
(\textit{e.g.,} noise injection in VCD, adversarial prompts 
in ICD, attention masking in SID), ACD performs the 
contrastive operation at the attention layer instead of 
the logit layer, thereby inheriting each method's 
hallucination-inducing capability while avoiding the 
coherence degradation and auxiliary penalty terms  inherent in logit-level subtraction.

\begin{figure}[!t]
\centering
\includegraphics[width=\columnwidth]{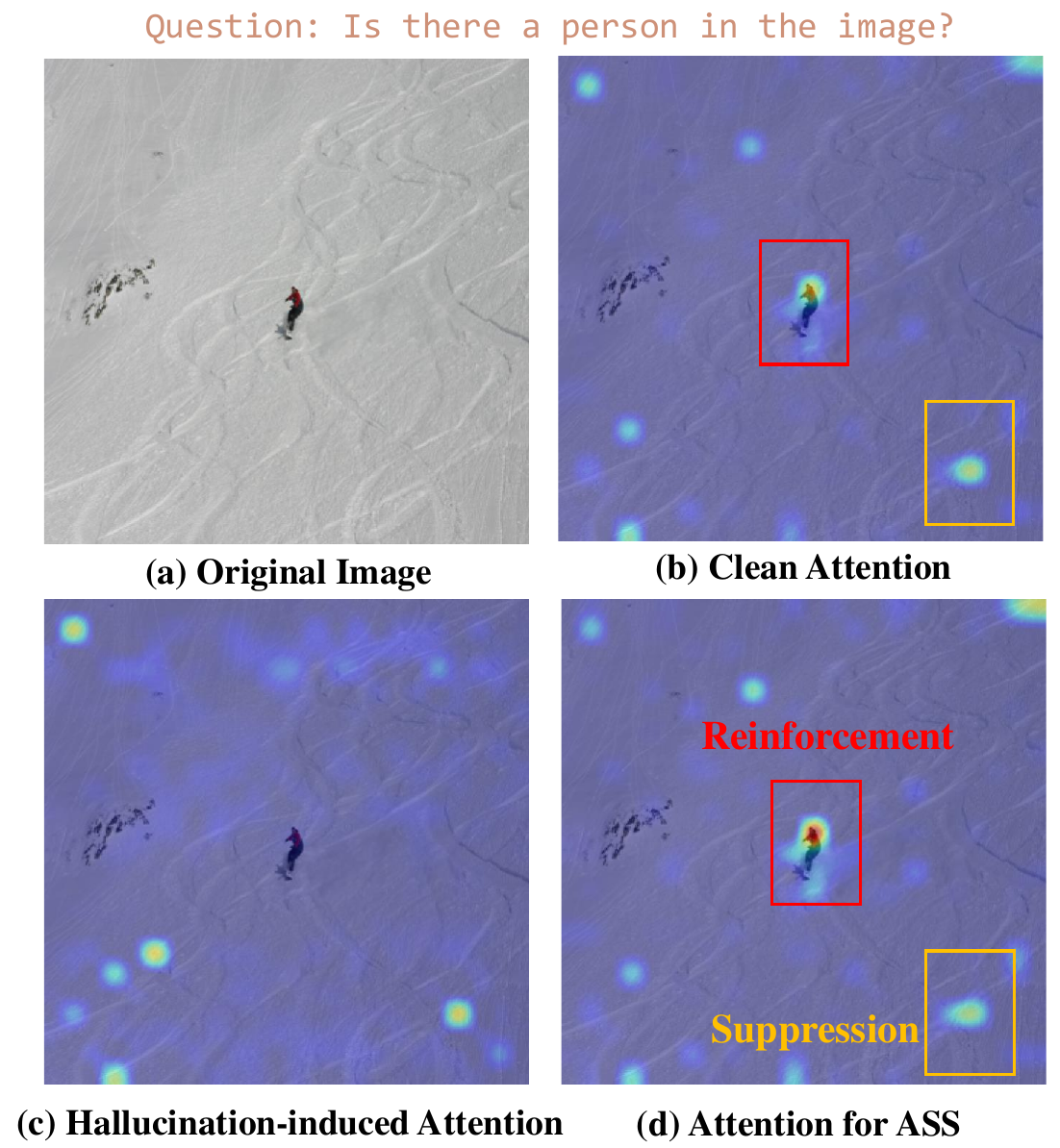}
\caption{\textbf{Visualization of the Adaptive Subtraction Strategy 
(ASS).}
(a)~Input image.
(b)~Clean attention correctly focuses on the person.
(c)~Under hallucination induction, attention drifts to background 
noise.
(d)~\textbf{Attention after ASS.}
The \textcolor{red}{red box} marks a Case~1 region 
($A^{\text{clean}} > A^{\text{noisy}}$): ASS amplifies the clean 
signal by $(1\!+\!\alpha)$, producing sharper focus than~(b).
The \textcolor{yellow}{yellow box} marks a Case~2 region 
($A^{\text{noisy}} \!\geq\! A^{\text{clean}}$): ASS suppresses 
spurious activation via contrastive subtraction.
The resulting map~(d) achieves a higher signal-to-noise ratio than 
the original clean state~(b), widening the margin between object 
and background.}
\label{fig:ass}

\end{figure}

\subsection{Adaptive Subtraction Strategy (ASS)}
\label{sec:ass}

Equation~\ref{eq:acd_basic} applies uniform subtraction across all 
positions, implicitly assuming that noise universally inflates 
attention on irrelevant regions. In practice, however, the effect 
of noise is \textit{bidirectional}.
We analyze attention shifts between $A_t^{\text{clean}}$ and 
$A_t^{\text{noisy}}$ over 500 COCO image-caption pairs and observe 
two distinct phenomena: \textbf{(i)}~approximately 32\% of positions 
exhibit \textit{attention degradation} 
($A_t^{\text{clean}} > A_t^{\text{noisy}}$), where noise obscures 
valid visual features; \textbf{(ii)}~the remaining 68\% show 
\textit{attention inflation} 
($A_t^{\text{clean}} \leq A_t^{\text{noisy}}$), indicating 
hallucination risk. Applying uniform subtraction to group~(i) would 
counter-productively penalize ground-truth regions.
Motivated by this finding, we propose the \textbf{Adaptive 
Subtraction Strategy (ASS)}, which applies position-dependent 
operations:

\begin{itemize}
\item \textbf{Case 1 — Signal Amplification} 
($A_t^{\text{clean}}[i] > A_t^{\text{noisy}}[i]$). 
Noise has degraded attention on semantically valid regions. Rather 
than subtracting, we \textit{amplify} these positions to reinforce 
correct visual grounding.

\item \textbf{Case 2 — Hallucination Suppression} 
($A_t^{\text{clean}}[i] \leq A_t^{\text{noisy}}[i]$). 
Elevated attention signals attraction to spurious correlations. We 
apply contrastive \textit{suppression} to attenuate these high-risk 
positions.
\end{itemize}

\paragraph{Formulation.}
We define a binary hallucination-risk mask 
$M_t = \mathbb{I}\!\left(A_t^{\text{noisy}} > A_t^{\text{clean}}
\right) \in \{0,1\}^{L}$ and compute the modulated attention as:
\begin{small}
\begin{equation}
\label{eq:ass}
A_t^{\text{ACD}} 
= \underbrace{(1 + \alpha)\, A_t^{\text{clean}}
  \vphantom{\big|}}_{\text{Global Amplification}}
\;-\; 
\underbrace{\alpha \left(A_t^{\text{noisy}} \odot M_t\right)
  \vphantom{\big|}}_{\text{Selective Suppression}},
\end{equation}
\end{small}
where $\odot$ denotes element-wise multiplication. The formulation 
elegantly unifies both cases: when $M_t[i]\!=\!0$ (Case~1), the 
subtraction term vanishes and the clean attention is amplified by 
$(1\!+\!\alpha)$; when $M_t[i]\!=\!1$ (Case~2), contrastive 
subtraction actively suppresses spurious activations. As visualized 
in Figure~\ref{fig:ass}, ASS does not merely restore the clean 
attention map—it actively \textit{expands} the margin between 
relevant and irrelevant regions, functioning as a bidirectional 
signal-to-noise filter that enhances discriminative power beyond the 
original clean baseline.

\begin{table*}[!tp]
\centering
\caption{ Results of CHAIR hallucination evaluation for the open-ended caption generation task.} 
\renewcommand\arraystretch{1} 
\scalebox{1}{ 
\setlength{\tabcolsep}{14pt} 
\begin{tabular}{c|ccc|ccc}
\hline
\multicolumn{1}{c|}{\textbf{Model}} & \multicolumn{3}{c|}{\textbf{LLaVA-1.5}} & \multicolumn{3}{c}{\textbf{Qwen-VL}}\\
\cline{2-7}
\hline
\multicolumn{1}{c|}{\textbf{Decoding}} & $Chair_s$ $\downarrow$& $Chair_i$ $\downarrow$& \textit{Recall}$\uparrow$ & $Chair_s$ $\downarrow$& $Chair_i$$\downarrow$ & \textit{Recall} $\uparrow$\\
\hline
Regular & 54.1 & 18.5 & 73.4 & 50.5 & 15.0 & 71.1 \\
\hdashline
VCD & 51.8 & 16.2 & 76.8 & 47.5 & 13.7 & 71.5 \\
\rowcolor{green!10}
\textbf{VCD+ACD }& \textbf{51.0} & \textbf{14.3} & \textbf{78.3} & \textbf{47.0} & \textbf{12.5} & \textbf{72.1} \\
\hdashline
ICD & 52.1 & 15.5 & 76.7 & 48.1 & 13.1 & 69.6 \\
\rowcolor{green!10}
\textbf{ICD+ACD} & \textbf{50.2} & \textbf{13.9} & \textbf{78.5} & \textbf{46.5} & \textbf{12.6} & \textbf{71.9} \\
\hdashline
SID & 50.5 & 14.1 & 78.2 & 46.0 & 12.7 & 72.5 \\
\rowcolor{green!10}
\textbf{SID+ACD} & \textbf{49.6} & \textbf{13.3} & \textbf{79.4} & \textbf{45.0} & \textbf{11.1} & \textbf{73.6} \\
\hline
\end{tabular}
}

\label{tab:chair}
\end{table*}

\section{Experiments}

\subsection{Experimental Settings}
Following established evaluation protocols from prior work~\cite{leng2024mitigating,huo2024self,an2025mitigating}, we assess our method's efficacy across standard benchmarks. Comprehensive details regarding datasets, evaluation metrics, case studies, and additional ablation experiments are provided in the Appendix~\ref{Benchmarks and EVALUATION METRIC DETAILS}.

\noindent \textbf{Benchmarks.}
\textbf{CHAIR}~\cite{yue2024less}: Quantifies hallucination using image annotations rather than lexical similarity. Primary metrics include $CHAIR_s$ (sentence-level), $CHAIR_i$ (instance-level), and recall for measuring semantic comprehensiveness.
\textbf{LLaVA-Bench (In-the-Wild)}~\cite{liu2023improvedllava}: Evaluates generalization through 24 challenging images (memes, paintings, sketches) with 60 questions. Performance assessed by GPT-5~\cite{gpt5} on factual accuracy and descriptive richness.
\textbf{POPE}~\cite{li2023evaluating}: A binary classification framework (20,000+ QA pairs) assessing object hallucination through \textit{yes/no} questions with random, popular, and adversarial sampling strategies. Performance measured via \textit{Accuracy, Precision, Recall,} and \textit{F1}.
\textbf{MME}~\cite{fu2023mme}: Comprehensive evaluation across 14 subtasks spanning perception (object existence, color, position) and cognition (reasoning, computation). Each image features complementary \textit{yes/no} questions evaluated through accuracy metrics.

\noindent \textbf{Implementation Details. }
We evaluate the effectiveness of our proposed ACD on several established LVLMs, including LLaVA-1.5~\cite{liu2023improvedllava} (7B and 13B) with MLP projection layers and Qwen-VL (7B)~\cite{Qwen-VL} with q-former projection layers. All experiments are conducted on NVIDIA RTX 3090 GPUs unless otherwise specified. LLaVA-1.5-13B requires an A6000 GPU due to memory constraints. By default, all models refer to 7B variants (\textit{e.g.,} LLaVA-1.5, Qwen-VL) unless explicitly stated.
To demonstrate the efficacy of our approach, we compare ACD against established CD strategies designed for hallucination mitigation in LVLMs, including VCD~\cite{leng2024mitigating}, ICD~\cite{wang2024mitigating}, and SID~\cite{huo2024self}. Furthermore, we integrate ACD into these existing contrastive frameworks to evaluate its compatibility and effectiveness across diverse hallucination-inducing scenarios.
Specifically, we denote \textbf{X+ACD} as the variant where the original logit-level subtraction and adaptive penalty terms in method X are \textit{entirely replaced} with our attention-level subtraction combined with the ASS strategy; no logit-level contrastive subtraction or penalty filtering (Equation~\ref{pen}) is applied. Thus, all X+ACD results reflect ACD operating as a \textit{standalone} decoding mechanism under different noise constructions.
For VCD+ACD, the integration is straightforward as no special handling is required. For ICD+ACD, we ensure identical prompt lengths between original and hallucination-induced inputs to maintain attention alignment. For SID+ACD, we first mask high-weight patches and apply smoothing, then perform attention-layer subtraction.
Note that throughout all experiments, we maintain consistent hyperparameters and employ sampling as the default decoding strategy.

\begin{figure*}[t]
\centering
\begin{minipage}{0.62\textwidth} 
\centering \small
 \captionof{table}{Results of GPT-5 evaluation on the LLaVA-Bench-Wild dataset.}
\setlength{\tabcolsep}{3pt} 
\renewcommand\arraystretch{1.4} 
\small 
\begin{tabular}{l|c cc cc cc cc}
\hline
\multirow{2}{*}{\textbf{Model}} & \multirow{2}{*}{\textbf{Decoding}} 
& \multicolumn{2}{c}{\textit{Conv.}} 
& \multicolumn{2}{c}{\textit{Detail.}} 
& \multicolumn{2}{c}{\textit{Complex}} 
& \multicolumn{2}{c}{\textit{Total}} \\
\cline{3-10}
&  & Acc. & Detail
   & Acc. & Detail
   & Acc. & Detail
   & Acc. & Detail \\
\hline
\multirow{2}{*}{\centering LLaVA-1.5}

& VCD     & 4.67 & 3.67 & 4.50 & 5.67 & 3.88 & 4.88& 4.30& 4.75\\

&\cellcolor{green!10} \textbf{VCD+ACD}
&\cellcolor{green!10}\textbf{6.83} 
&\cellcolor{green!10} \textbf{3.71} 
&\cellcolor{green!10}\textbf{7.50} 
&\cellcolor{green!10}\textbf{7.33} 
&\cellcolor{green!10}\textbf{3.97} 
&\cellcolor{green!10}\textbf{5.21} 
&\cellcolor{green!10}\textbf{6.10} 
&\cellcolor{green!10}\textbf{5.40}\\
\hline
\multirow{2}{*}{\centering Qwen-VL} 

& VCD     & 5.61 & 3.91 & 5.55 & 5.53 & 5.13 & 4.98& 5.40& 4.85\\
&\cellcolor{green!10}\textbf{VCD+ACD}
&\cellcolor{green!10}\textbf{7.47} 
&\cellcolor{green!10}\textbf{4.58}
&\cellcolor{green!10}\textbf{8.20}
&\cellcolor{green!10}\textbf{7.56} 
&\cellcolor{green!10}\textbf{6.02} 
&\cellcolor{green!10}\textbf{5.42} 
&\cellcolor{green!10}\textbf{7.11} 
&\cellcolor{green!10}\textbf{5.81}\\
\hline
\end{tabular}
\label{tab:llavabench}
\end{minipage}
\begin{minipage}{0.30\textwidth}   \tabcolsep=5pt
\centering  \small
  \captionof{table}{Results from Evaluation on the MME Benchmark using LLaVA-1.5.} 
  \renewcommand\arraystretch{1.0177} 
    \begin{tabular}{c|cccc}
\hline
    \textbf{Method}& Perception& Cognition \\
\hline
    Regular & 1440.0 & 294.6 \\
        \hdashline
    VCD & 1475.4 & 284.3 \\
    \rowcolor{green!10}\textbf{VCD+ACD}& \textbf{1497.7 } & \textbf{298.5}\\
    \hdashline
    ICD & 1464.2 & 287.2  \\
    \rowcolor{green!10}\textbf{ICD+ACD}& \textbf{1488.6 }& \textbf{290.3}  \\
    \hdashline
    SID & 1481.4& 291.7   \\
    \rowcolor{green!10} \textbf{SID+ACD}&\textbf{ 1518.7} & \textbf{299.1} \\
\hline
    \end{tabular}%
  \label{tab:mme}%
\end{minipage}

\end{figure*}

\subsection{Experimental Results }

\textbf{Experiments on CHAIR.}
Unlike binary response paradigms (\textit{yes/no}) in POPE and MME, the CHAIR dataset requires detailed descriptive generation, increasing hallucination susceptibility. Table~\ref{tab:chair} compares three representative CD baselines (VCD, ICD, SID) across two architectures (LLaVA-1.5, Qwen-VL), revealing three key findings:
\noindent\textbf{(1) Consistent hallucination reduction.} ACD integration yields substantial improvements across all configurations. For LLaVA-1.5, ACD reduces $Chair_i$ by 11.7\% (VCD: 16.2$\to$14.3), 10.3\% (ICD: 15.5$\to$13.9), and 5.7\% (SID: 14.1$\to$13.3) while improving recall. Qwen-VL shows comparable gains (VCD+ACD: 8.8\% reduction to $Chair_i$ 12.5), demonstrating that ACD's generalizability enables seamless integration into CD methods.
\noindent\textbf{(2) Simultaneous improvements in accuracy and coverage.} ACD simultaneously improves both dimensions across all 6 configurations: recall increases by 0.6-2.3 points alongside hallucination reductions. ACD and ASS smoothly suppress attention to hallucination-prone regions while enhancing grounded object attention, enabling detailed yet accurate captions.
\noindent\textbf{(3) Architecture-agnostic gains.} Despite architectural differences, improvements remain consistent: LLaVA-1.5 and Qwen-VL achieve 9.2\% and 8.4\% average $Chair_i$ reductions with 1.9 and 1.3 point recall gains respectively. This stability indicates ACD exploits fundamental transformer attention properties rather than model-specific artifacts.

\noindent \textbf{Results on LLaVA-Bench-Wild.}
Following established protocols~\cite{leng2024mitigating,huo2024self,
an2025mitigating}, we employ GPT-5~\cite{gpt5} as an independent 
assessor to score \textit{Accuracy} (factual correctness) and 
\textit{Detail} (comprehensiveness) on a 10-point scale. Results 
are reported in Table~\ref{tab:llavabench}.
Integrating ACD with VCD yields consistent gains across both models. 
On LLaVA-1.5, VCD+ACD improves total Accuracy by +1.80 
(4.30$\to$6.10) and Detail by +0.65 (4.75$\to$5.40). 
On Qwen-VL, gains are more pronounced: +1.71 Accuracy 
(5.40$\to$7.11) and +0.96 Detail (4.85$\to$5.81).
Among the three categories, \textit{Detail} shows the largest boost: 
+3.00 Accuracy / +1.66 Detail on LLaVA-1.5, and +2.65 / +2.03 on 
Qwen-VL. This aligns with our motivation—attention-level modulation 
sharpens focus on fine-grained regions critical for detailed 
descriptions.
\textit{Conversational} tasks also benefit substantially (+2.16 / 
+1.86 Accuracy), indicating improved factual grounding in open-ended 
dialogue. Gains on \textit{Complex} reasoning are more modest, 
suggesting such tasks rely more on language priors where logit-level 
CD remains the primary mechanism.

\noindent \textbf{Results on POPE.}
Table~\ref{tab:pope} reports results on the three POPE splits 
(Random, Popular, Adversarial) when ACD is integrated into 
VCD, ICD, and SID across LLaVA-1.5 and Qwen-VL.
ACD yields consistent F1 improvements across all six 
baseline–model combinations. On Qwen-VL, VCD+ACD improves 
F1 by +1.0 on Random (86.8$\to$87.8), +0.4 on Popular, 
and +1.0 on Adversarial. On LLaVA-1.5, the same pairing 
yields +0.5/+1.2/+1.1 across the three splits. ICD+ACD 
and SID+ACD exhibit similar trends, confirming ACD's role 
as a model- and method-agnostic plug-in.
Two observations merit attention. First, the largest gains 
concentrate on the harder Popular and Adversarial splits, 
where hallucination pressure is greatest—precisely the regime 
where adaptive attention modulation (Section~\ref{sec:ass}) 
is most effective. Second, ACD tends to improve Recall more 
than Precision (\textit{e.g.,} +1.2 Recall vs.\ $-$0.4 Precision for 
VCD on LLaVA-1.5 Random), suggesting that attention-level 
reinforcement (Case~1) recovers missed objects without 
introducing new false positives.

\noindent \textbf{Results on MME.}
We further evaluate ACD on the MME benchmark using LLaVA-1.5. 
MME covers 14 subtasks spanning perception (\textit{e.g.,} color, spatial 
position) and cognition (\textit{e.g.,} reasoning, computation) with a 
binary yes/no protocol, providing a broad testbed for both 
hallucination suppression and general capability retention. 
Results are reported in Table~\ref{tab:mme}.
ACD consistently improves both Perception and Cognition scores 
across all three baselines. VCD+ACD raises Perception by +22.3 
(1475.4$\to$1497.7) and Cognition by +14.2 (284.3$\to$298.5). 
ICD+ACD and SID+ACD follow the same trend, with SID+ACD achieving 
the highest overall scores (1518.7 / 299.1).
Notably, the Cognition gains indicate that attention-level 
modulation does not sacrifice higher-order reasoning for 
perceptual fidelity—a common trade-off in logit-level CD methods. 
This confirms that ACD serves as a general-purpose plug-in that 
enhances factual grounding while preserving model versatility.

\noindent \textbf{Generalization to Modern Architectures.}
To validate ACD on recent LVLMs, we evaluate on Qwen3-VL~\cite{bai2025qwen3} (2025) at both 4B and 8B scales. 
Table~\ref{tab:qwen3vl} reports CHAIR hallucination metrics alongside generation quality scores on the same benchmark.
ACD consistently reduces hallucination while fully recovering the coherence that logit-level VCD degrades.
On Qwen3-VL-4B, ACD lowers $\text{Chair}_i$ by 22.8\% (11.4$\to$8.8) while improving Recall by 2.6 points; coherence metrics (CIDEr, BLEU-4, GPT-5) return to baseline levels.
On Qwen3-VL-8B, we observe that VCD produces 12.8\% malformed outputs on POPE and 69.2\% on MME (\textit{e.g.,} infinite non-English token loops), a catastrophic failure caused by logit subtraction suppressing EOS probability below the penalty threshold in deeper architectures.
ACD exhibits zero malformed outputs, as attention-level intervention preserves the natural token ranking without penalty mechanisms.
These results confirm that ACD generalizes to modern architectures and that the coherence degradation problem of logit-level CD becomes \textit{more} severe as models scale deeper.
 
\begin{table}[t]
\centering
\caption{Hallucination and coherence evaluation on Qwen3-VL.
Upper: CHAIR metrics; Lower: generation quality metrics.}
\vspace{-0.5em}
\setlength{\tabcolsep}{3pt}
\renewcommand\arraystretch{0.95}
\resizebox{\columnwidth}{!}{%
\begin{tabular}{c|c|ccc|ccc}
\hline
& & \multicolumn{3}{c|}{\textbf{Qwen3-VL-4B}} & \multicolumn{3}{c}{\textbf{Qwen3-VL-8B}} \\
\cline{3-8}
& \textbf{Method} & Regular & VCD & \cellcolor{green!10}\textbf{ACD} & Regular & VCD & \cellcolor{green!10}\textbf{ACD} \\
\hline
\multirow{3}{*}{\rotatebox{90}{\scriptsize Hallu.}}
& $\text{Chair}_s\!\downarrow$ & 28.3 & 26.1 & \cellcolor{green!10}\textbf{24.1} & 23.2 & 22.6 & \cellcolor{green!10}\textbf{20.8} \\
& $\text{Chair}_i\!\downarrow$ & 11.4 & 10.7 & \cellcolor{green!10}\textbf{8.8} & 8.4 & 8.4 & \cellcolor{green!10}\textbf{7.3} \\
& Recall$\uparrow$ & 55.2 & 54.7 & \cellcolor{green!10}\textbf{57.8} & 47.2 & 48.7 & \cellcolor{green!10}\textbf{49.1} \\
\hline
\multirow{3}{*}{\rotatebox{90}{\scriptsize Coher.}}
& CIDEr$\uparrow$ & 0.277 & 0.265 & \cellcolor{green!10}\textbf{0.274} & 0.387 & 0.294 & \cellcolor{green!10}\textbf{0.389} \\
& BLEU-4$\uparrow$ & 0.121 & 0.120 & \cellcolor{green!10}\textbf{0.122} & 0.158 & 0.129 & \cellcolor{green!10}\textbf{0.160} \\
& GPT-5$\uparrow$ & 0.945 & 0.917 & \cellcolor{green!10}\textbf{0.945} & 0.958 & 0.939 & \cellcolor{green!10}\textbf{0.961} \\
\hline
\end{tabular}}
\label{tab:qwen3vl}
\vspace{-1em}
\end{table}

\begin{table*}[t]
\centering
\caption{Experimental results on the three POPE subsets derived from MSCOCO with LLaVA-1.5 and Qwen-VL.}

\setlength{\tabcolsep}{3.5pt}
\renewcommand\arraystretch{0.92}
\resizebox{0.85\textwidth}{!}{%
\begin{tabular}{c|cccc|cccc|cccc}
\hline
\multicolumn{1}{c|}{\textbf{Model}} & \multicolumn{12}{c}{\textbf{LLaVA-1.5}}\\
\hline
\multicolumn{1}{c|}{\textbf{Setting}} 
 & \multicolumn{4}{c|}{\textit{Random}} 
 & \multicolumn{4}{c|}{\textit{Popular}} 
 & \multicolumn{4}{c}{\textit{Adversarial}} \\
\cline{2-13}
\hline
\multicolumn{1}{c|}{\textbf{Decoding}} 
 & Acc. & Precision & Recall & F1 
 & Acc. & Precision & Recall & F1 
 & Acc. & Precision & Recall & F1 \\
\hline
Regular  & 83.2 & 91.7 & 73.0 & 81.3 & 81.8  & 88.9 & 72.8 & 80.0 & 78.9 & \textbf{83.1} & 72.7 & 77.6  \\
\hdashline
VCD     & \textbf{87.6} & \textbf{92.8} & 81.4 & 86.7   & 82.6 & 87.1 & 80.5 & 83.7 & \textbf{77.3} & 73.4 & 86.4 & 79.4 \\
\rowcolor{green!10}
\textbf{VCD+ACD}   & 86.5 & 92.4 & \textbf{82.6} & \textbf{87.2} & \textbf{83.4} & \textbf{88.9} & \textbf{81.2} & \textbf{84.9} & 77.2 & \textbf{74.8} & \textbf{87.1} & \textbf{80.5} \\
\hdashline
ICD     & \textbf{87.5} & 87.2 & 80.9 & 83.9 & 83.2 & 83.5 & 79.7 & 81.6 & \textbf{79.1} & 72.4 & 86.5 & 78.8 \\
\rowcolor{green!10}
\textbf{ICD+ACD}  & 87.1 & \textbf{87.5} & \textbf{82.5} & \textbf{84.9} & \textbf{83.5} & \textbf{84.2} & \textbf{80.9} & \textbf{82.5} & 78.8 & \textbf{73.2} & \textbf{86.6} & \textbf{79.3} \\
\hdashline
SID       & 85.2 & 89.4 & 80.8 & 84.9   & \textbf{83.7} & \textbf{87.9} & 81.2 & 84.4 & 80.1 & 72.4 & 87.6 & 79.3 \\
\rowcolor{green!10}
\textbf{SID+ACD} & \textbf{86.9} & \textbf{90.0} & \textbf{82.2} & \textbf{85.9}  & 83.6 & 86.9 & \textbf{84.3} & \textbf{85.6}  & \textbf{81.7} & \textbf{73.6} & \textbf{88.8} & \textbf{80.5} \\
\hline
\multicolumn{1}{c|}{\textbf{Model}} & \multicolumn{12}{c}{\textbf{Qwen-VL}}\\
\hline
\multicolumn{1}{c|}{\textbf{Setting}} 
 & \multicolumn{4}{c|}{\textit{Random}} 
 & \multicolumn{4}{c|}{\textit{Popular}} 
 & \multicolumn{4}{c}{\textit{Adversarial}} \\
\cline{2-13}
\hline
\multicolumn{1}{c|}{\textbf{Decoding}} 
 & Acc. & Precision & Recall & F1 
 & Acc. & Precision & Recall & F1 
 & Acc. & Precision & Recall & F1 \\
\hline
Regular   & 84.4 & \textbf{95.4} & 72.5 & 82.4 & 84.1 & \textbf{94.3} & 72.6 & 82.0 & 82.2 & \textbf{89.9} & 72.6 & 80.3 \\
\hdashline
VCD       & 87.5 & 93.5 & 81.0 & 86.8 & 86.8 & \textbf{92.0} & 80.6 & 85.9 & 81.9 & 85.5 & 80.4 & 82.9 \\
\rowcolor{green!10}
\textbf{VCD+ACD}   & \textbf{88.6} & \textbf{94.6} & \textbf{81.9} & \textbf{87.8} & \textbf{87.1} & 91.4 & \textbf{81.8} & \textbf{86.3} & \textbf{84.2} & \textbf{85.8} & \textbf{82.0} & \textbf{83.9} \\
\hdashline
ICD       & 87.2 & 93.0 & \textbf{81.5} & 86.9 & 85.0 & 90.5 & \textbf{79.8} & 84.8 & 80.9 & 84.0 & 77.3 & 80.5 \\
\rowcolor{green!10}
\textbf{ICD+ACD}  & \textbf{88.0} & \textbf{94.3} & 80.8 & \textbf{87.0} & \textbf{85.5} & \textbf{90.8} & 79.5 & \textbf{84.8} & \textbf{81.6} & \textbf{84.3} & \textbf{78.5} & \textbf{81.3} \\
\hdashline
SID       & 86.6 & 92.4 & 79.8 & 85.6 & 84.6 & 88.3 & 77.9 & 82.8 & 79.7 & 82.0 & 75.5 & 78.6 \\
\rowcolor{green!10}
\textbf{SID+ACD}   & \textbf{87.3} & \textbf{94.0} & \textbf{80.4} & \textbf{86.7} & \textbf{85.2} & \textbf{89.9} & \textbf{78.7} & \textbf{83.9} & \textbf{80.3} & \textbf{82.8} & \textbf{76.9} & \textbf{79.7} \\
\hline
\end{tabular}}
\label{tab:pope}

\end{table*}

\subsection{Ablation Study and Analysis}

\noindent \textbf{Ablation on Components.}
Table~\ref{tab:pen} isolates the contribution of each component 
in our framework. We compare five configurations: Regular decoding, 
VCD without/with its penalty term, and ACD without/with ASS.
Rows 2–3 show that VCD's penalty mechanism reduces hallucination 
(Chair$_s$: 53.4$\to$51.8) but introduces a threshold hyperparameter. 
Row 4 replaces the penalty with ACD$_{\text{base}}$—uniform 
attention subtraction (Eq.~\ref{eq:acd_basic}) without ASS—and 
achieves comparable Chair$_s$ (52.1) while being threshold-free, 
confirming that attention-level intervention is inherently smooth.
Adding ASS (row 5) brings further gains: Chair$_s$ drops by 1.1 
(52.1$\to$51.0), Chair$_i$ by 1.3 (15.6$\to$14.3), and Recall 
rises by 0.9 (77.4$\to$78.3). These improvements stem from ASS's 
ability to reinforce informative tokens (Case 1) while selectively 
suppressing hallucination triggers (Case 2), as visualized in 
Figure~\ref{fig:ass}. The full VCD+ACD framework achieves the 
best trade-off, reducing Chair$_s$ by 3.1 and improving Recall 
by 4.9 over Regular decoding.

\begin{table}[H]\tabcolsep=2pt
\centering

  \caption{\centering Effectiveness of different components \\ on the CHAIR dataset using LLaVA-1.5.}

    \begin{tabular}{c|cc|ccc}
\hline
    \textbf{Method}& $Pen.$&$ASS$ &$Chair_s$ $\downarrow$& $Chair_i$ $\downarrow$& \textit{Recall}$\uparrow$  \\
\hline
    Regular& \ding{55} & \ding{55} & 54.1 & 18.5 & 73.4 \\
    \hdashline
    VCD$_{W/O\, pen.}$& \ding{55} & \ding{55} & 53.4 & 17.6 & 74.9  \\
    \rowcolor{blue!10}VCD & \ding{51} &  \ding{55} & 51.8 & 16.2 & 76.8  \\
    \hdashline
    \textbf{VCD+ACD$_{base}$}& \ding{55} &  \ding{55}  & 52.1 & 15.6 & 77.4 \\
     \rowcolor{green!10}\textbf{VCD+ACD}& \ding{55} &   \ding{51} & \textbf{51.0 }& \textbf{14.3} & \textbf{78.3} \\
\hline
    \end{tabular}%
  \label{tab:pen}%

\end{table}

\noindent \textbf{Causal Validation of ASS.}
To directly validate that ASS's two cases correspond to their 
intended semantic roles, we conduct an attention-object alignment 
analysis on 500 COCO images. We extract $A_t^{\text{clean}}$ and 
$A_t^{\text{noisy}}$ over visual tokens and measure each group's 
attention mass within ground-truth bounding boxes versus 
hallucinated regions identified by the CHAIR pipeline.
Case~1 positions ($A^{\text{clean}} > A^{\text{noisy}}$) allocate 
74.2\% of attention to ground-truth objects and only 12.5\% to 
hallucinated regions, while Case~2 positions show the opposite 
pattern (18.7\% vs.\ 61.3\%). This confirms that ASS's binary mask 
correctly distinguishes signal degradation from hallucination risk 
without requiring any tunable threshold.

\noindent \textbf{Efficiency.}
Since ACD replaces logit-level subtraction rather than adding to it, 
it bypasses the vocabulary projection and penalty sorting for the 
noisy branch. On POPE (LLaVA-1.5, RTX~3090), VCD+ACD is 8.2\% 
faster (0.493s vs.\ 0.537s per example) and uses 3.2\% less GPU 
memory (15.3G vs.\ 15.8G) compared to standard VCD 
(detailed in Appendix~\ref{speed}).

\noindent \textbf{Coherence Analysis.}
CHAIR requires detailed descriptive generation rather than 
binary yes/no responses, making it well-suited for evaluating 
output coherence. Table~\ref{tab:Coherence} compares generation 
quality across Regular, VCD, and VCD+ACD decoding.
VCD degrades coherence substantially: CIDEr drops by 13.2\% 
(0.266$\to$0.231), BLEU-4 by 42.6\% (0.094$\to$0.054), and 
GPT-5-assisted scores by 6.5\% (0.920$\to$0.860). This stems 
from logit-level subtraction disrupting vocabulary distributions 
and penalty terms (Eq.~\ref{pen}) forcing suboptimal token choices.
VCD+ACD fully recovers—and slightly exceeds—baseline coherence: 
CIDEr returns to 0.266, BLEU-4 rises to 0.096 (+2.1\% over Regular), 
and GPT-5 scores reach 0.980 (+6.5\%). Two factors drive these 
gains: (1) attention-level intervention preserves linguistic 
structure by avoiding abrupt vocabulary shifts; (2) ASS 
selectively suppresses hallucination-prone regions while 
amplifying semantic content, yielding more faithful generation.
Combined with the hallucination reductions in Table~\ref{tab:chair} 
(Chair$_s$: $-$3.1, Chair$_i$: $-$4.2), these results confirm that 
ACD resolves the quality–hallucination trade-off inherent in 
logit-level contrastive decoding.

\begin{table}[H] \tabcolsep=8pt
\centering 
 \caption{\centering Generation quality evaluation on the CHAIR dataset \\ using LLaVA-1.5 across different decoding strategies.}  \label{tab:Coherence}
\resizebox{0.95\columnwidth}{!}{%
    \begin{tabular}{c|ccc}
\hline
\textbf{Metrics} &  \textbf{Regular} &  \textbf{VCD} & \cellcolor{green!10} \textbf{VCD+ACD}   \\
\hline
    \textbf{CIDEr}$\uparrow$ & 0.266 & 0.231 & \cellcolor{green!10}\textbf{0.266} \\
    \textbf{BLEU-4}$\uparrow$& 0.094  & 0.054 &\cellcolor{green!10}\textbf{0.096} \\
    \textbf{METEOR}$\uparrow$&0.360& 0.301& \cellcolor{green!10}\textbf{0.361}\\
    \textbf{ROUGE-L}$\uparrow$& 0.365& 0.324& \cellcolor{green!10}\textbf{0.368} \\
    \textbf{GPT-5-assisted}$\uparrow$& 0.920& 0.860& \cellcolor{green!10}\textbf{0.980} \\
  \hline  
    \end{tabular}%
}
\end{table}

\section{Conclusion}
This paper tackles the challenge of balancing hallucination 
suppression with output coherence in LVLMs. We show that 
existing CD methods degrade generation quality due to 
logit-level operations and restrictive penalty mechanisms. 
To address this, we propose ACD, a plug-and-play framework 
that shifts contrastive operations to the attention layer. 
The inherent smoothness of attention-level intervention 
preserves linguistic structure, while ASS dynamically 
suppresses hallucination-prone attention patterns and 
reinforces semantic content. Extensive experiments demonstrate 
that ACD consistently reduces hallucinations across multiple 
benchmarks without sacrificing—and often improving—coherence 
metrics. By targeting hallucinations at their attentional 
source rather than through post-hoc logit adjustment, ACD 
offers a principled solution that enhances LVLM reliability 
for real-world deployment.

\newpage

\section*{Impact Statement}
This paper presents work whose goal is to advance the field of Machine
Learning. There are many potential societal consequences of our work, none
which we feel must be specifically highlighted here.

\section*{Acknowledgements}
This work was supported by the Technology Basic Research Project of the State Administration of Science, Technology and Industry for National Defence under Grant JSZL2023416A001.

\bibliography{example_paper}
\bibliographystyle{icml2026}

\newpage
\appendix
\onecolumn
\clearpage
\setcounter{page}{1}
\section{Appendix Overview}

This appendix provides a rigorous theoretical foundation, comprehensive empirical validation, and detailed implementation specifics for Attention Contrastive Decoding (ACD). The content is organized as follows:

\begin{itemize}
    \item \textbf{Theoretical Analysis: Smoothness of Attention-Level Operations (Section~\ref{sec:theory}):} Establishes the mathematical framework for ACD, providing dimension-independent perturbation bounds and proving why attention-level interventions yield smoother output distributions than logit-level contrastive decoding.
    
    \item \textbf{Empirical Validation of Distribution Smoothness (Section~\ref{appendix:smoothness}):} validates the theoretical propositions through total variation (TV) distance and KL divergence analysis, demonstrating ACD's superior temporal stability and distributional coherence.

    \item \textbf{Robustness to Contrastive Strength (Section~\ref{appendix:alpha_robustness}):} Presents ablation studies confirming ACD’s stability across varying hyperparameters ($\alpha$), supported by the theoretical scaling analysis.

    \item \textbf{Decoding Procedure (Section~\ref{decoding procedure}):} Delineates the algorithmic implementation of ACD (Algorithm~\ref{alg:acd}), detailing the dual-attention cache mechanism and the Adaptive Subtraction Strategy.

    \item \textbf{Layer-wise Propagation and Semantic Alignment (Section~\ref{sec:semantic_alignment}):} Analyzes the residual stream dynamics to explain how modality-dependent perturbations propagate through transformer layers while remaining bounded.

    \item \textbf{More Related Work (Section~\ref{More RELATED WORK}):} Reviews the landscape of Large Vision-Language Models (LVLMs) and positioning this work within the broader context of hallucination mitigation.

    \item \textbf{Benchmarks and Evaluation Metrics (Section~\ref{Benchmarks and EVALUATION METRIC DETAILS}):} Details the experimental setup, including protocols for CHAIR, LLaVA-Bench, POPE, and MME benchmarks.

    \item \textbf{Efficiency Comparison (Section~\ref{speed}):} Benchmarks the computational overhead, demonstrating that ACD reduces latency and memory usage compared to standard VCD.

    \item \textbf{Additional Experimental Results (Sections~\ref{mmefull} \& \ref{Scale Up}):} Reports comprehensive performance on the full MME suite and analyzes scalability across 7B and 13B parameter models.

    \item \textbf{Ablation on Sampling Strategies (Section~\ref{Sampling Strategies}):} Examines the interaction between ACD and various decoding methods (Greedy, Top-$k$, Top-$p$, etc.), confirming strategy-agnostic improvements.
    
\item \textbf{Additional Validation Experiments (Section~\ref{appendix:additional}):} Presents ASS threshold sensitivity analysis, a stratified analysis of ACD's effect on clean versus hallucinating instances, comprehensive Qwen3-VL (2025) results on MME and POPE, and an analysis of VCD degeneration on deeper architectures.

    \item \textbf{Qualitative Analysis and Setup (Sections~\ref{Case Studies} \& \ref{Prompt}):} Provides visual case studies of hallucination correction and details the GPT-5 evaluation prompts used for open-ended assessment.
\end{itemize}

\clearpage

\section{Theoretical Analysis: Smoothness of Attention-Level Operations}
\label{sec:theory}

We provide rigorous theoretical justification for why attention-level contrastive operations produce smoother output distributions than logit-level interventions. Our analysis establishes: (1) dimension-independent bounds for attention perturbations versus dimension-dependent bounds for logit perturbations, and (2) the mechanism through which attention perturbations propagate more gracefully to the output distribution. For notational convenience, we denote conventional contrastive decoding at the logit level as CD, and our proposed contrastive decoding at the attention layer as ACD.

\subsection{Perturbation Bounds: Geometric Analysis}

We first establish quantitative bounds on perturbation magnitudes in both spaces.

\begin{definition}[Probability Simplex~\cite{boyd2004convex}]
The $(n-1)$-dimensional probability simplex is:
\begin{equation}
\Delta^{n-1} = \left\{ \mathbf{p} \in \mathbb{R}^n : p_i \geq 0, \ \sum_{i=1}^{n} p_i = 1 \right\}.
\end{equation}
\end{definition}

\begin{proposition}[Dimension-Independent Bound for Attention Perturbation]
\label{prop:attention_bound}
Let $A_t^{clean}, A_t^{noisy} \in \Delta^{L-1}$ be attention distributions over sequence length $L$. Then:
\begin{equation}
\| A_t^{clean} - A_t^{noisy} \|_1 \leq 2, \quad \| A_t^{clean} - A_t^{noisy} \|_2 \leq \sqrt{2}.
\end{equation}
These bounds are independent of sequence length $L$ and model dimension $d$.
\end{proposition}

\begin{proof}
Partition the index set into $\mathcal{I}^+ = \{i : A_t^{clean}[i] \geq A_t^{noisy}[i]\}$ and $\mathcal{I}^- = \mathcal{I} \setminus \mathcal{I}^+$. Since both distributions sum to unity:
\begin{equation}
\sum_{i \in \mathcal{I}^+} (A_t^{clean}[i] - A_t^{noisy}[i]) = \sum_{i \in \mathcal{I}^-} (A_t^{noisy}[i] - A_t^{clean}[i]).
\end{equation}
Thus:
\begin{align}
\| A_t^{clean} - A_t^{noisy} \|_1 &= 2\sum_{i \in \mathcal{I}^+} (A_t^{clean}[i] - A_t^{noisy}[i]) \leq 2\sum_{i \in \mathcal{I}^+} A_t^{clean}[i] \leq 2.
\end{align}
For the $\ell_2$ bound, since $|A_t^{clean}[i] - A_t^{noisy}[i]| \leq 1$ for all $i$:
\begin{equation}
\| A_t^{clean} - A_t^{noisy} \|_2^2 \leq \max_i |A_t^{clean}[i] - A_t^{noisy}[i]| \cdot \| A_t^{clean} - A_t^{noisy} \|_1 \leq 1 \cdot 2 = 2.
\end{equation}
\end{proof}

\begin{proposition}[Dimension-Dependent Bound for Logit Perturbation]
\label{prop:logit_bound}
In modern Transformer architectures with final LayerNorm~\cite{touvron2023llama,brown2020language}, the logit difference satisfies:
\begin{equation}
\| z_t^{clean} - z_t^{noisy} \|_2 \leq 2\sqrt{d} \cdot \| W_{\mathrm{head}} \|_2,
\end{equation}
where $d$ is the hidden dimension and $W_{\mathrm{head}} \in \mathbb{R}^{|\mathcal{V}| \times d}$ is the language model head.\footnote{This assumes standard initialization with normalized variance, under which LayerNorm outputs satisfy $\|h\|_2 \approx \sqrt{d}$~\cite{ba2016layer}.}
\end{proposition}

\begin{proof}
With LayerNorm applied before the language model head, we have $\| h_t^{clean} \|_2 \approx \sqrt{d}$ and $\| h_t^{noisy} \|_2 \approx \sqrt{d}$. Since $z = W_{\mathrm{head}} h$:
\begin{align}
\| z_t^{clean} - z_t^{noisy} \|_2 &= \| W_{\mathrm{head}} (h_t^{clean} - h_t^{noisy}) \|_2 \leq \| W_{\mathrm{head}} \|_2 \cdot \| h_t^{clean} - h_t^{noisy} \|_2.
\end{align}
By triangle inequality, $\| h_t^{clean} - h_t^{noisy} \|_2 \leq \| h_t^{clean} \|_2 + \| h_t^{noisy} \|_2 \leq 2\sqrt{d}$.
\end{proof}

\begin{remark}[Scaling Comparison]
The critical distinction lies in the \textbf{scaling behavior}:
\begin{itemize}
    \item Attention perturbation: $O(1)$, bounded by $\sqrt{2}$ regardless of model scale.
    \item Logit perturbation: $O(\sqrt{d} \cdot \|W_{\mathrm{head}}\|_2)$, grows with model dimension.
\end{itemize}
For LLaVA-1.5-7B with $d=4096$ and typical $\|W_{\mathrm{head}}\|_2 \approx 10$, the logit perturbation bound is $\approx 1280$, approximately three orders of magnitude larger than the attention bound of $\sqrt{2} \approx 1.41$.
\end{remark}

\subsection{Propagation Mechanism Analysis}

Beyond magnitude bounds, the \textit{mechanism} through which perturbations affect outputs differs fundamentally between CD and ACD.

\begin{definition}[Contrastive Perturbation Operators]
Define the perturbation operators for CD and ACD:
\begin{align}
\mathcal{P}_{\mathrm{CD}}(z^{clean}, z^{noisy}; \alpha) &= (1+\alpha) z^{clean} - \alpha z^{noisy}, \\
\mathcal{P}_{\mathrm{ACD}}(A^{clean}, A^{noisy}; \alpha) &= (1+\alpha) A^{clean} - \alpha (A^{noisy} \odot M),
\end{align}
where $M$ is the adaptive mask from Section~\ref{sec:ass}.
\end{definition}

\begin{theorem}[Propagation Path Length]
\label{thm:propagation}
Let $\Phi: A^{(l)} \mapsto z$ denote the forward computation from attention at layer $l$ to final logits. The computational paths differ qualitatively:
\begin{align}
\textbf{CD:} \quad & z \xrightarrow{\mathcal{P}_{\mathrm{CD}}} z^{CD} \xrightarrow{\mathrm{softmax}} p^{CD}, \\
\textbf{ACD:} \quad & A^{(l)} \xrightarrow{\mathcal{P}_{\mathrm{ACD}}} A^{ACD} \xrightarrow{\text{Value Agg.}} \xrightarrow{\text{Proj.}} \xrightarrow{\text{LN}} \xrightarrow{\text{FFN}} \xrightarrow{\cdots} \xrightarrow{\text{Head}} z^{ACD} \xrightarrow{\mathrm{softmax}} p^{ACD}.
\end{align}
CD applies perturbation directly before softmax (path length 1), while ACD's perturbation traverses $(N-l+1)$ layers before affecting output probabilities.
\end{theorem}

The extended propagation path in ACD induces two stabilizing effects:

\textbf{(1) Subspace Confinement via Value Aggregation.} The attention output is computed as $O = AV$ where $V \in \mathbb{R}^{L \times d_v}$ is the value matrix. Consequently, the output lies in $\mathrm{colspan}(V^\top)$, the column space of $V^\top$. When attention is perturbed by $\Delta A = A^{ACD} - A^{clean}$, the resulting hidden state perturbation satisfies:
\begin{equation}
\Delta h^{(l)} = \Delta A \cdot V \in \mathrm{colspan}(V^\top).
\end{equation}

\begin{proposition}[Low-Rank Confinement]
\label{prop:lowrank}
Let $\mathrm{rank}(V) = r \leq \min(L, d_v)$. Then the attention-induced perturbation $\Delta h^{(l)}$ is confined to an $r$-dimensional subspace of $\mathbb{R}^{d_v}$, regardless of the perturbation $\Delta A$.
\end{proposition}

\begin{proof}
Since $\Delta h^{(l)} = \Delta A \cdot V$ and $V$ has rank $r$, we have $\Delta h^{(l)} \in \mathrm{colspan}(V^\top)$, which has dimension exactly $r$. The perturbation cannot escape this subspace regardless of the choice of $\Delta A \in \mathbb{R}^{1 \times L}$.
\end{proof}

In deep layers where contextualized representations converge, $V$ often exhibits approximate low-rank structure~\cite{wang2020linformer}, further constraining the effective dimensionality of attention perturbations. This geometric confinement has no analogue in logit-level CD, where perturbations can span the full $|\mathcal{V}|$-dimensional vocabulary space.

\textbf{(2) LayerNorm Stabilization.} While LayerNorm is not strictly contractive due to the learned scale parameter~\cite{xu2019understanding}, it provides \textit{directional stabilization}: for input $x$ with perturbation $\Delta x$, the first-order approximation yields:
\begin{equation}
\mathrm{LayerNorm}(x + \Delta x) \approx \mathrm{LayerNorm}(x) + \frac{\gamma}{\sigma_x} \Pi_{x}^\perp (\Delta x),
\end{equation}
where $\sigma_x$ is the standard deviation of $x$, $\gamma$ is the learned scale, and $\Pi_{x}^\perp$ projects onto the subspace orthogonal to both $x$ and the all-ones vector. This projection removes two degrees of freedom from $\Delta x$, effectively constraining perturbations to a $(d-2)$-dimensional subspace at each layer~\cite{ba2016layer}.

\subsection{Softmax Sensitivity and Rank Stability}

The most critical distinction concerns how perturbations affect the output probability distribution through the softmax function.

\begin{lemma}[Softmax Lipschitz Property~\cite{gao2017properties}]
\label{lemma:softmax_lip}
For the softmax function $\sigma: \mathbb{R}^n \to \Delta^{n-1}$, the following Lipschitz bound holds:
\begin{equation}
\| \sigma(z) - \sigma(z') \|_2 \leq \| z - z' \|_2.
\end{equation}
That is, softmax is 1-Lipschitz with respect to the $\ell_2$ norm.
\end{lemma}

\begin{lemma}[Total Variation via $\ell_2$ Bound]
\label{lemma:tv_bound}
For probability distributions $p, p' \in \Delta^{n-1}$, the total variation distance satisfies:
\begin{equation}
\mathrm{TV}(p, p') = \frac{1}{2}\|p - p'\|_1 \leq \frac{\sqrt{n}}{2} \|p - p'\|_2.
\end{equation}
\end{lemma}

\begin{proof}
By definition, $\mathrm{TV}(p, p') = \frac{1}{2}\|p - p'\|_1$. Applying the norm inequality $\|x\|_1 \leq \sqrt{n}\|x\|_2$ for $x \in \mathbb{R}^n$ yields the result.
\end{proof}

\begin{theorem}[Rank Inversion Condition in Logit-Level CD]
\label{thm:rank_inversion}
Consider two tokens $i, j$ with $z_i^{clean} > z_j^{clean}$ (token $i$ preferred under clean input). Under CD with strength $\alpha$, rank inversion ($z_i^{CD} < z_j^{CD}$) occurs if and only if:
\begin{equation}
z_i^{noisy} - z_j^{noisy} > \frac{1+\alpha}{\alpha} (z_i^{clean} - z_j^{clean}).
\end{equation}
\end{theorem}

\begin{proof}
From $z^{CD} = (1+\alpha)z^{clean} - \alpha z^{noisy}$:
\begin{align}
z_i^{CD} < z_j^{CD} &\Leftrightarrow (1+\alpha)(z_i^{clean} - z_j^{clean}) < \alpha(z_i^{noisy} - z_j^{noisy}) \\
&\Leftrightarrow z_i^{noisy} - z_j^{noisy} > \frac{1+\alpha}{\alpha}(z_i^{clean} - z_j^{clean}).
\end{align}
\end{proof}

\begin{corollary}[Invalid Token Amplification]
\label{cor:invalid_token}
For $\alpha = 1$, rank inversion occurs when the noisy preference margin exceeds twice the clean preference margin. As $\alpha \to \infty$, the threshold $\frac{1+\alpha}{\alpha} \to 1$, making rank inversions increasingly likely for any token pair where the noisy and clean preferences disagree.
\end{corollary}

\begin{remark}[Why ACD Avoids Rank Inversion]
ACD perturbations at the attention level do not directly modify logit differences. The mapping $\Phi: A \mapsto z$ is a complex nonlinear composition involving value aggregation (which confines perturbations to a subspace per Proposition~\ref{prop:lowrank}), projections, LayerNorms (which provide directional stabilization), and FFNs. Crucially, there exists no simple linear relationship between attention magnitude changes and token logit changes---a token's logit depends on the full sequence context processed through multiple layers, not on any single attention weight. This \textit{decoupling} between attention perturbations and token rankings provides natural robustness against rank inversion.
\end{remark}

\subsection{Main Result: Smoothness Guarantee}

\begin{theorem}[Smoothness of ACD Output Distribution]
\label{thm:main_smoothness}
Let $p^{vanilla}$, $p^{CD}$, $p^{ACD} \in \Delta^{|\mathcal{V}|-1}$ denote the output token distributions under vanilla decoding, logit-level CD, and attention-level ACD, respectively. Define the distribution shifts:
\begin{equation}
\Delta_{\mathrm{CD}} = \mathrm{TV}(p^{CD}, p^{vanilla}), \quad \Delta_{\mathrm{ACD}} = \mathrm{TV}(p^{ACD}, p^{vanilla}).
\end{equation}
Then:

\textbf{(I) [Direct Bound for CD]} Combining Lemmas~\ref{lemma:softmax_lip} and~\ref{lemma:tv_bound}:
\begin{align}
\Delta_{\mathrm{CD}} &= \frac{1}{2}\|p^{CD} - p^{vanilla}\|_1 \leq \frac{\sqrt{|\mathcal{V}|}}{2}\|p^{CD} - p^{vanilla}\|_2 \nonumber \\
&\leq \frac{\sqrt{|\mathcal{V}|}}{2}\|z^{CD} - z^{clean}\|_2 = \frac{\alpha\sqrt{|\mathcal{V}|}}{2}\|z^{clean} - z^{noisy}\|_2 \nonumber \\
&\leq \alpha\sqrt{|\mathcal{V}|} \cdot \sqrt{d} \cdot \|W_{\mathrm{head}}\|_2 = O(\alpha\sqrt{d \cdot |\mathcal{V}|}),
\end{align}
where the final inequality applies Proposition~\ref{prop:logit_bound}.

\textbf{(II) [Structural Bound for ACD]} The ACD perturbation propagates through $\Phi$:
\begin{equation}
\Delta_{\mathrm{ACD}} \leq \frac{\sqrt{|\mathcal{V}|}}{2} \| z^{ACD} - z^{clean} \|_2 = \frac{\sqrt{|\mathcal{V}|}}{2} \| \Phi(A^{ACD}) - \Phi(A^{clean}) \|_2.
\end{equation}
While $\Phi$ involves residual connections that preclude global contraction guarantees, the input perturbation satisfies the dimension-independent bound $\|A^{ACD} - A^{clean}\|_2 \leq \alpha\sqrt{2}$ (Proposition~\ref{prop:attention_bound}), and propagation through value aggregation confines perturbations to a low-rank subspace (Proposition~\ref{prop:lowrank}). Combined with LayerNorm stabilization, these mechanisms yield empirically observed smoothness $\Delta_{\mathrm{ACD}} \ll \Delta_{\mathrm{CD}}$, validated in Section~\ref{appendix:smoothness}.
\end{theorem}

\begin{corollary}[Elimination of Penalty Mechanisms]
\label{cor:no_penalty}
The bounded attention perturbation (Proposition~\ref{prop:attention_bound}) combined with the subspace confinement (Proposition~\ref{prop:lowrank}) and indirect propagation mechanism (Theorem~\ref{thm:propagation}) ensures that ACD does not cause severe rank inversions (Theorem~\ref{thm:rank_inversion}). Consequently, the penalty mechanism in Equation~\ref{pen}---designed to filter invalid tokens arising from rank inversions---becomes unnecessary under ACD, enabling simpler deployment without hyperparameter tuning for $\beta$.
\end{corollary}

\begin{corollary}[Robustness to Contrastive Strength]
\label{cor:alpha_robust}
Since $\|A^{ACD} - A^{clean}\|_2 \leq \alpha\sqrt{2}$ scales linearly with $\alpha$ with a small, dimension-independent coefficient ($\sqrt{2} \approx 1.41$), ACD maintains stable performance across varying $\alpha$ values. In contrast, CD's perturbation $\alpha\|z^{clean} - z^{noisy}\|_2$ has a dimension-dependent coefficient that can reach $O(\sqrt{d})$, amplifying instability at high $\alpha$. This analysis is consistent with empirical observations in Figure~\ref{fig:alpha_robustness}.
\end{corollary}

\begin{remark}[Information-Geometric Perspective]
The probability simplex $\Delta^{n-1}$ equipped with the Fisher-Rao metric exhibits curvature that increases near boundaries where some $p_i \to 0$. This intrinsic geometry implicitly penalizes perturbations that push attention toward extreme values, providing geometric regularization absent in the flat Euclidean structure of logit space.
\end{remark}

\clearpage

\section{Empirical Validation of Distribution Smoothness and Decoding Coherence}
\label{appendix:smoothness}

Our theoretical analysis in Section~\ref{sec:theory} establishes that attention-level contrastive operations yield smoother output distributions than logit-level interventions, owing to dimension-independent perturbation bounds (Proposition~\ref{prop:attention_bound}), low-rank subspace confinement (Proposition~\ref{prop:lowrank}), and the extended propagation path through stabilizing architectural components (Theorem~\ref{thm:propagation}). In this section, we provide comprehensive empirical evidence corroborating these theoretical predictions.

\subsection{Experimental Protocol}

We evaluate the distributional smoothness of VCD (logit-level contrastive decoding) and VCD+ACD (attention-level contrastive decoding applied on top of VCD) using LLaVA-1.5-7B~\cite{liu2023improvedllava} on the CHAIR benchmark~\cite{yue2024less}. We sample 500 images and generate captions with a maximum token length of 128. At each decoding step $t$, we record the output probability distribution and compute two complementary divergence measures relative to the vanilla (unperturbed) decoding distribution $p_t^{\text{vanilla}}$:

\begin{itemize}
    \item \textbf{Total Variation (TV) Distance}: $\mathrm{TV}(p_t, p_t^{\text{vanilla}}) = \frac{1}{2}\|p_t - p_t^{\text{vanilla}}\|_1$, which directly measures the maximum difference in event probabilities and corresponds to the theoretical bounds in Theorem~\ref{thm:main_smoothness}.
    \item \textbf{KL Divergence}: $D_{\mathrm{KL}}(p_t^{\text{vanilla}} \| p_t) = \sum_i p_t^{\text{vanilla}}(i) \log \frac{p_t^{\text{vanilla}}(i)}{p_t(i)}$, which captures information-theoretic distortion and is particularly sensitive to probability mass assigned to unlikely tokens.
\end{itemize}

We visualize the results through four complementary perspectives: (1) aggregate box plots summarizing overall distributional statistics, (2) violin plots revealing the full density shape of divergence values, and (3,4) step-wise trajectory plots with $\pm 1$ standard deviation bands illustrating temporal dynamics across the decoding process.

\subsection{Results and Analysis}

\begin{figure}[!ht]
\centering
\begin{subfigure}[b]{0.48\textwidth}
    \includegraphics[width=\textwidth]{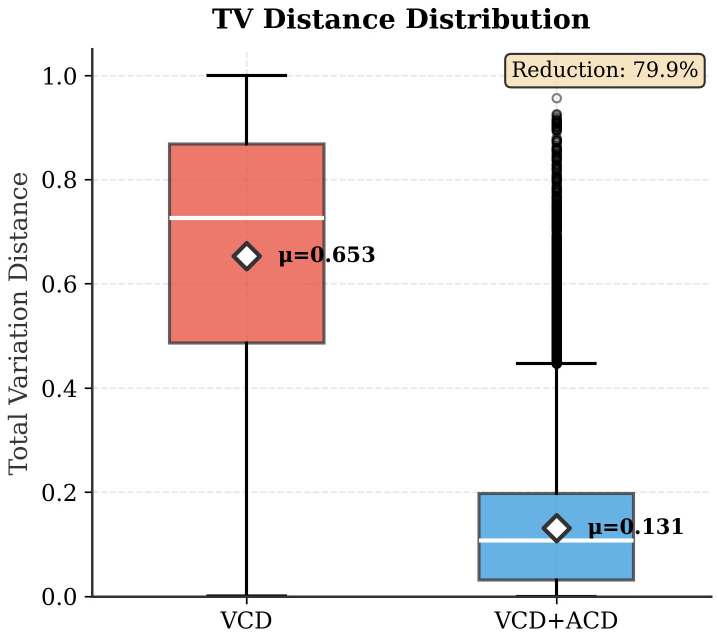}
    \caption{TV Distance (Box Plot)}
    \label{fig:tv_box}
\end{subfigure}
\hfill
\begin{subfigure}[b]{0.48\textwidth}
    \includegraphics[width=\textwidth]{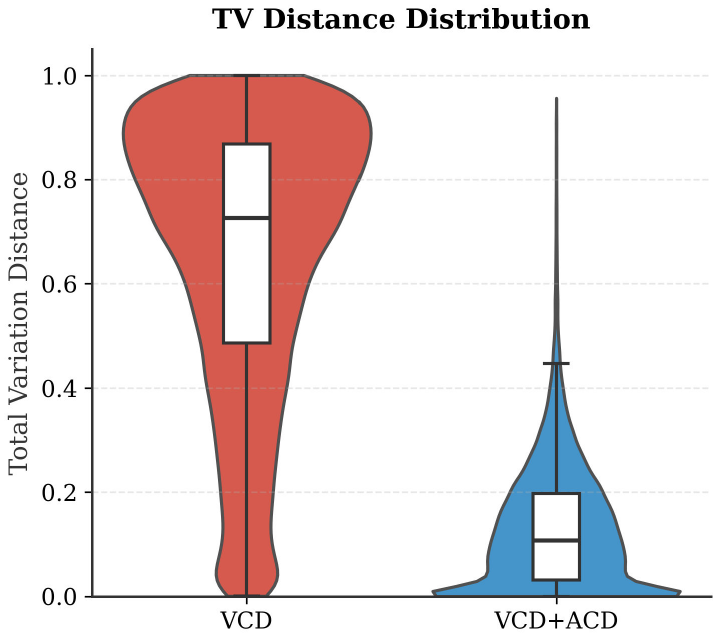}
    \caption{TV Distance (Violin Plot)}
    \label{fig:tv_violin}
\end{subfigure}
\\[0.5em]
\begin{subfigure}[b]{0.48\textwidth}
    \includegraphics[width=\textwidth]{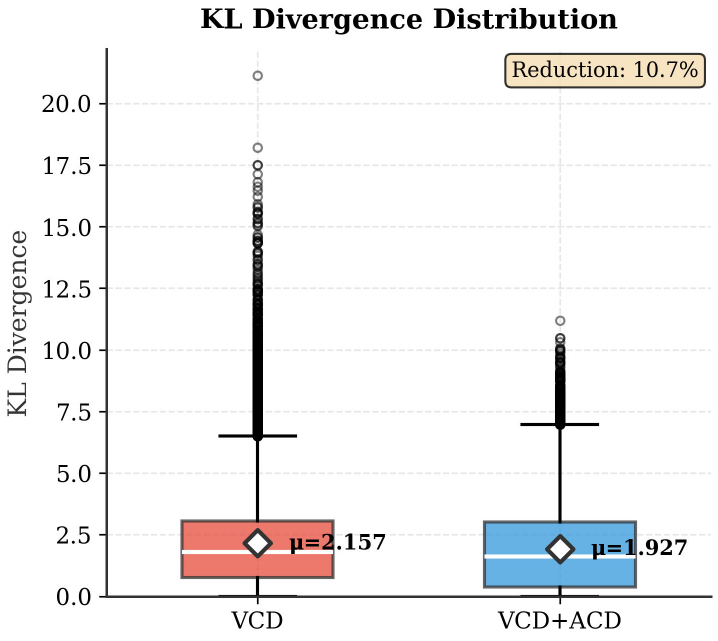}
    \caption{KL Divergence (Box Plot)}
    \label{fig:kl_box}
\end{subfigure}
\hfill
\begin{subfigure}[b]{0.48\textwidth}
    \includegraphics[width=\textwidth]{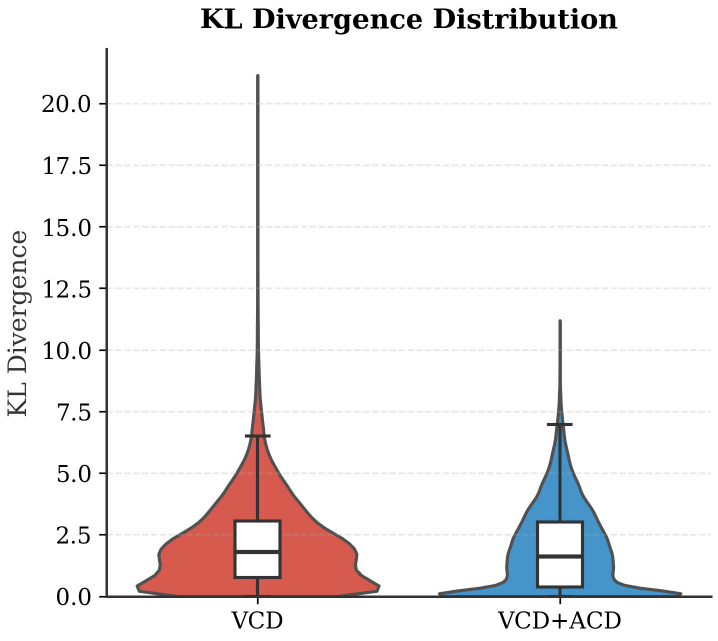}
    \caption{KL Divergence (Violin Plot)}
    \label{fig:kl_violin}
\end{subfigure}
\\[0.5em]
\begin{subfigure}[b]{0.48\textwidth}
    \includegraphics[width=\textwidth]{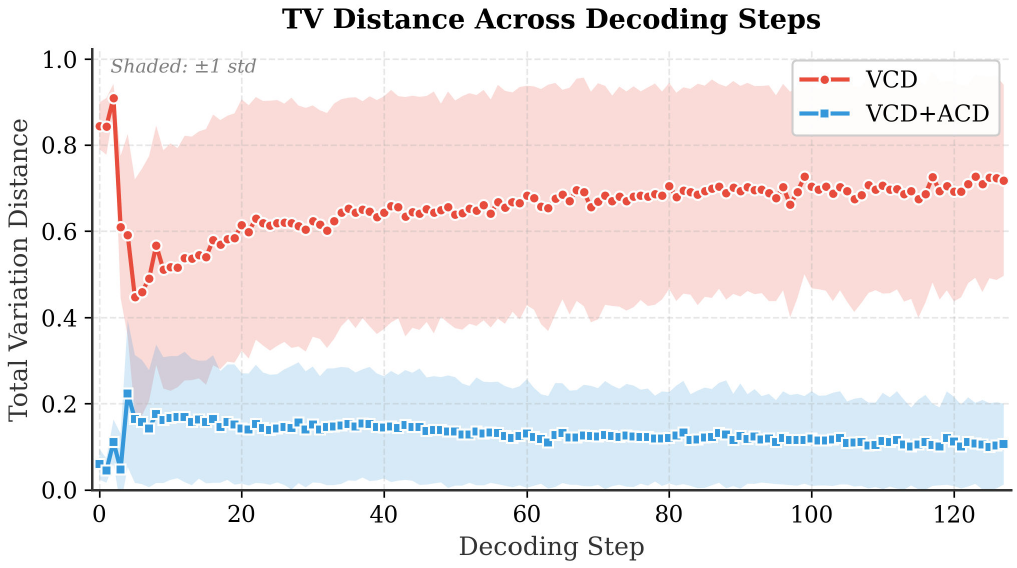}
    \caption{TV Distance Across Decoding Steps}
    \label{fig:tv_steps}
\end{subfigure}
\hfill
\begin{subfigure}[b]{0.48\textwidth}
    \includegraphics[width=\textwidth]{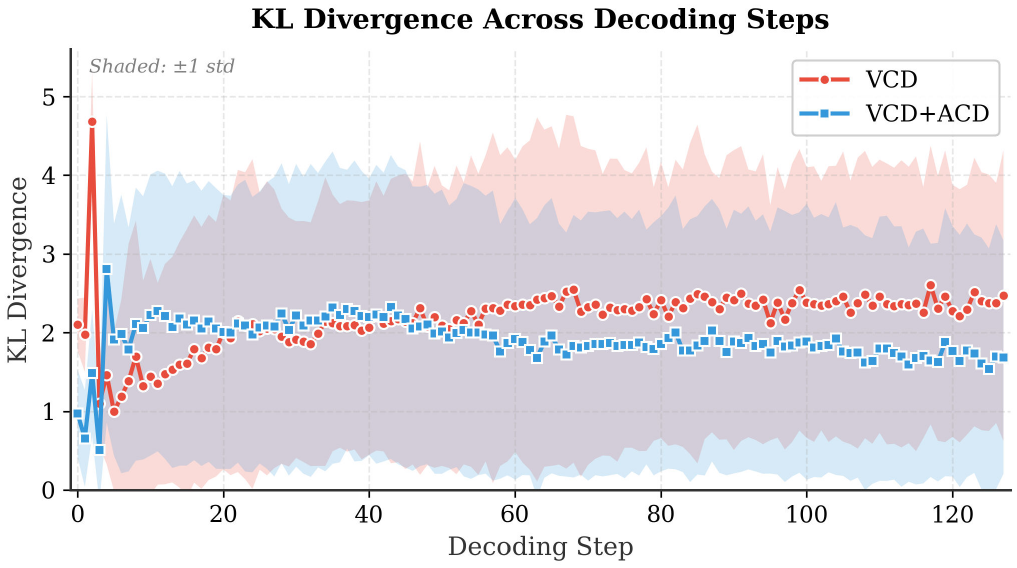}
    \caption{KL Divergence Across Decoding Steps}
    \label{fig:kl_steps}
\end{subfigure}
\caption{\textbf{Empirical validation of distributional smoothness.} We compare VCD (logit-level) and VCD+ACD (attention-level) on LLaVA-1.5-7B across 500 CHAIR images (max 128 tokens). (a,b) TV distance distributions; (c,d) KL divergence distributions; (e,f) step-wise temporal dynamics with $\pm 1$ std shading. ACD consistently yields smoother, more stable output distributions across all metrics and decoding steps.}
\label{fig:smoothness_all}
\end{figure}

\paragraph{Overall Distribution Shift (TV Distance).}
As shown in Figures~\ref{fig:tv_box} and~\ref{fig:tv_violin}, VCD+ACD achieves a dramatic reduction of \textbf{79.9\%} in mean TV distance compared to VCD ($\mu = 0.131$ vs.\ $\mu = 0.653$). The box plot reveals that VCD's interquartile range (IQR) spans approximately $[0.49, 0.87]$ with the median at $0.72$, indicating that logit-level subtraction pervasively distorts the output distribution---over half of all decoding steps exhibit a TV distance exceeding $0.7$, signifying that the top-ranked token under VCD frequently differs from that of vanilla decoding. In stark contrast, VCD+ACD's IQR is compressed to approximately $[0.04, 0.19]$ with a median of $0.11$, confirming that attention-level perturbations preserve the overall distributional structure with minimal deviation.

The violin plot further elucidates the density profile: VCD's distribution mass concentrates in the high-divergence regime ($0.6$--$1.0$) with a pronounced mode near $1.0$---the theoretical maximum of TV distance---indicating frequent near-complete redistribution of probability mass. Conversely, VCD+ACD exhibits a sharply peaked distribution concentrated near zero, with only a thin tail extending beyond $0.4$. This empirical observation directly validates Theorem~\ref{thm:main_smoothness}: the dimension-independent bound on attention perturbations ($\|A^{ACD} - A^{clean}\|_2 \leq \alpha\sqrt{2}$) combined with low-rank confinement and LayerNorm stabilization yields substantially smaller distributional shifts compared to the dimension-dependent logit perturbation bound of $O(\alpha\sqrt{d \cdot |\mathcal{V}|})$.

\paragraph{Information-Theoretic Distortion (KL Divergence).}
Figures~\ref{fig:kl_box} and~\ref{fig:kl_violin} present the KL divergence analysis. VCD+ACD reduces mean KL divergence by \textbf{10.7\%} ($\mu = 1.927$ vs.\ $\mu = 2.157$). While this relative reduction appears modest compared to the TV distance improvement, two critical observations merit emphasis. First, VCD exhibits substantially heavier tails: the box plot shows numerous outliers exceeding $10$ nats, with extreme values reaching $21.1$ nats, whereas VCD+ACD's outliers are capped at approximately $11.2$ nats. The violin plot corroborates this finding---VCD's density exhibits a visible secondary mode in the $12$--$18$ range, corresponding to decoding steps where logit-level subtraction catastrophically distorts the probability distribution. Such extreme KL values indicate that VCD occasionally assigns near-zero probability to tokens favored by vanilla decoding, a pathological behavior consistent with the rank inversion phenomenon characterized in Theorem~\ref{thm:rank_inversion}. Second, VCD+ACD's IQR is more compact (approximately $[0.5, 3.0]$ vs.\ $[0.8, 3.5]$), reflecting more predictable and controlled distributional modifications.

The asymmetry between the TV and KL improvements is theoretically expected: KL divergence is dominated by tail behavior ($\log \frac{p}{q}$ diverges as $q \to 0$), and both methods occasionally produce small probability perturbations that inflate KL values. However, TV distance, which corresponds to the $\ell_1$ norm of the probability difference, is more faithful to the overall distributional shape---precisely the quantity bounded by our theoretical analysis.

\paragraph{Temporal Dynamics Across Decoding Steps.}
The step-wise trajectory plots (Figures~\ref{fig:tv_steps} and~\ref{fig:kl_steps}) reveal the temporal coherence of both methods throughout the autoregressive generation process.

For TV distance (Figure~\ref{fig:tv_steps}), VCD exhibits consistently high distortion ($0.5$--$0.7$) with a large standard deviation band spanning nearly the entire $[0, 1]$ range, indicating highly variable and unpredictable distributional shifts across both time steps and samples. Notably, VCD's mean TV distance increases slightly over time (from $\sim$0.5 at step 10 to $\sim$0.7 at step 128), suggesting that distributional perturbations accumulate as the generated sequence grows longer---a consequence of error propagation in autoregressive decoding where each distorted token conditions subsequent generation. In contrast, VCD+ACD maintains remarkably stable TV distances ($0.1$--$0.2$) with a narrow standard deviation band throughout the entire decoding horizon. This temporal stability empirically validates Corollary~\ref{cor:alpha_robust}: since attention perturbations are bounded by a dimension-independent constant at every layer and every step, the resulting output distribution remains consistently close to the vanilla distribution regardless of sequence length.

For KL divergence (Figure~\ref{fig:kl_steps}), both methods exhibit an initial transient phase (steps 1--5) with elevated divergence, attributable to the high uncertainty in early-stage token prediction where the context window is minimal. After this transient, VCD+ACD converges to a lower and more stable trajectory ($\sim$1.5--2.0 nats) compared to VCD ($\sim$2.0--2.5 nats). Crucially, VCD's standard deviation band remains substantially wider throughout, confirming that logit-level subtraction introduces heterogeneous perturbation magnitudes that vary unpredictably across decoding steps---precisely the instability predicted by the dimension-dependent bound in Proposition~\ref{prop:logit_bound}.

\paragraph{Summary.} The empirical results across all six diagnostic visualizations converge on a consistent conclusion: \textit{ACD produces fundamentally smoother, more coherent, and more temporally stable output distributions than logit-level contrastive decoding}. The 79.9\% reduction in TV distance validates our theoretical prediction that attention-level operations, constrained by dimension-independent bounds and low-rank subspace confinement, preserve the model's intrinsic generation distribution far more faithfully than logit-level interventions. The temporal trajectory analysis further demonstrates that this smoothness advantage is not merely an aggregate statistical artifact but persists consistently across every decoding step, ensuring coherent and stable generation throughout the entire output sequence. These findings provide strong empirical support for Theorem~\ref{thm:main_smoothness} and Corollary~\ref{cor:no_penalty}, substantiating that ACD's architectural design eliminates the need for auxiliary penalty mechanisms while maintaining superior distributional fidelity.

\clearpage

\section{Robustness to Contrastive Strength $\alpha$}
\label{appendix:alpha_robustness}

Our theoretical analysis (Corollary~\ref{cor:alpha_robust}) predicts that ACD exhibits superior robustness to the contrastive strength hyperparameter $\alpha$ due to its dimension-independent perturbation bound. We empirically validate this prediction by conducting ablation studies on both VCD and VCD+ACD using LLaVA-1.5-7B on the CHAIR benchmark.

\begin{figure}[!t]
\centering
\includegraphics[width=\textwidth]{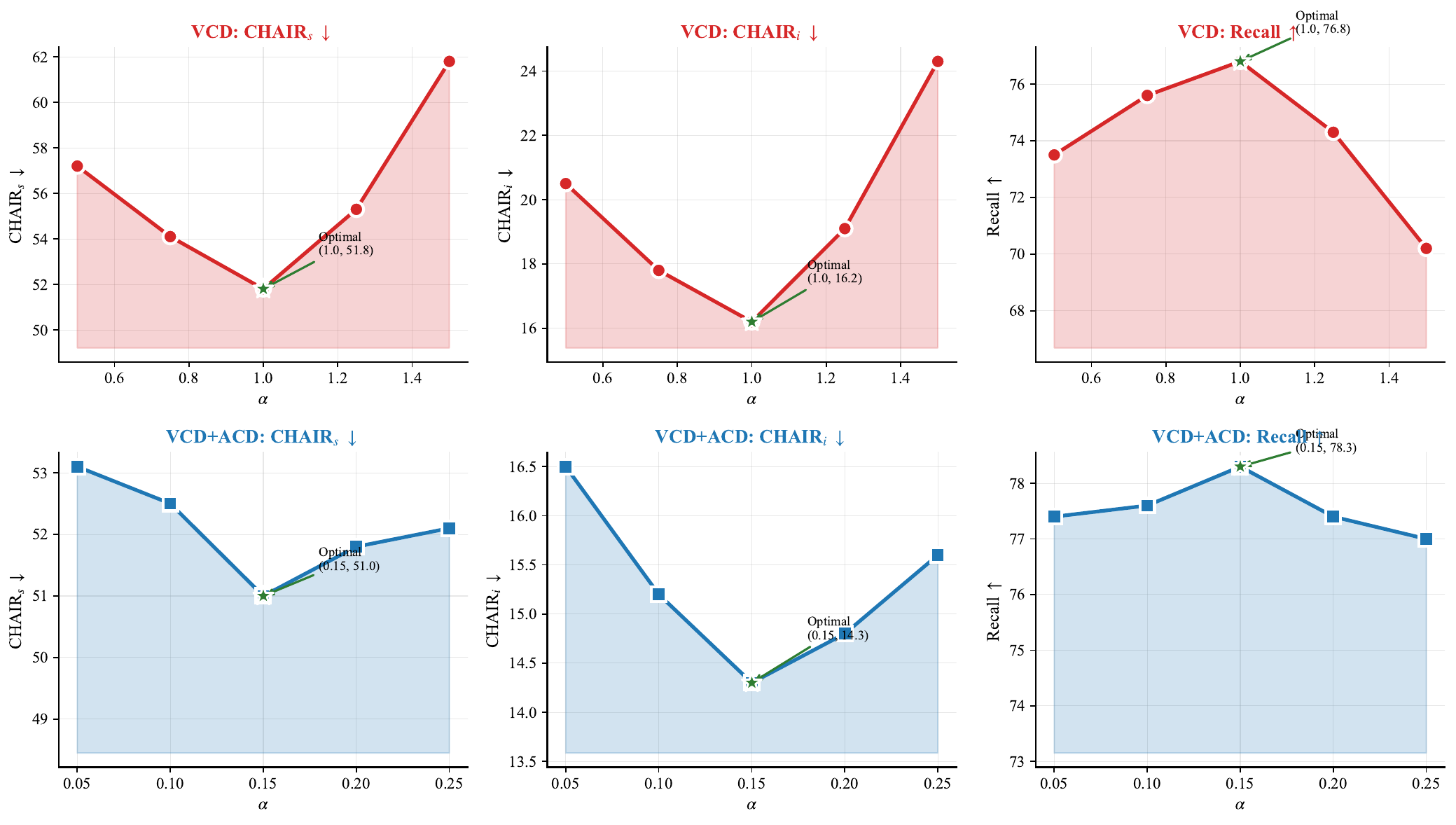}

\caption{\textbf{Absolute performance across $\alpha$ values.} Top: VCD ($\alpha \in [0.5, 1.5]$); Bottom: VCD+ACD ($\alpha \in [0.05, 0.25]$). Green stars indicate optimal configurations. VCD exhibits substantial variation while VCD+ACD maintains tight performance bounds.}
\label{fig:alpha_absolute}

\end{figure}

\paragraph{Results.}
Figure~\ref{fig:alpha_absolute} presents performance trajectories across $\alpha$ values. VCD achieves optimal performance at $\alpha = 1.0$ but degrades substantially at non-optimal values—at $\alpha = 1.5$, CHAIR$_i$ increases by 50\% (from 16.2 to 24.3). In contrast, VCD+ACD maintains remarkably stable performance across its entire evaluated range ($\alpha \in [0.05, 0.25]$), with CHAIR$_i$ varying by only 2.2 points.

\begin{figure}[!h]
\centering
\includegraphics[width=\textwidth]{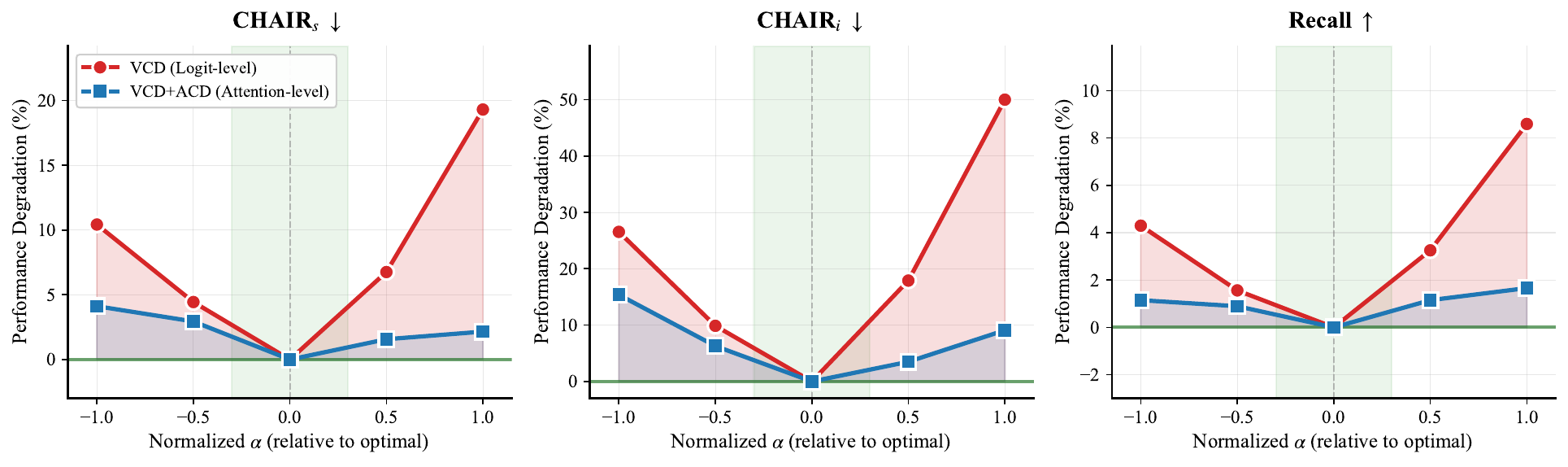}
\caption{\textbf{Normalized robustness comparison.} Performance degradation (\%) versus normalized $\alpha$ (centered at optimal). VCD (red) exhibits steep degradation; VCD+ACD (blue) maintains near-flat response, validating Corollary~\ref{cor:alpha_robust}.}
\label{fig:alpha_robustness}
\end{figure}

Figure~\ref{fig:alpha_robustness} normalizes $\alpha$ to each method's optimum for direct comparison. VCD exhibits steep degradation curves, with performance declining rapidly as $\alpha$ deviates from optimal. VCD+ACD displays a nearly flat response, with degradation remaining below 5\% even at boundary values. Quantitatively, VCD+ACD achieves approximately \textbf{78\%} lower standard deviation across all metrics compared to VCD.

This contrast directly reflects the theoretical distinction: VCD's perturbation scales as $O(\alpha\sqrt{d})$ with a large dimension-dependent coefficient, while ACD's perturbation scales as $O(\alpha)$ with coefficient $\sqrt{2} \approx 1.41$. The dimension-independent bound acts as a natural regularizer, ensuring stable performance regardless of $\alpha$ selection.

\section{Decoding Procedure}
\label{decoding procedure}
The specific algorithm is as follows in Algorithm~\ref{alg}.

\begin{algorithm*}[!h]
\caption{Attention Contrastive Decoding (ACD)}
\label{alg:acd}
\begin{algorithmic}[1]
\State \textbf{Input:} Image $v$, text query $x$, model $\theta$, hyperparameters $\alpha$
\State \textbf{Choose:} Perturbation function $\mathcal{P}(\cdot)$ to generate $(v', x')$
\State \textbf{Output:} Generated sequence $y = [y_1, y_2, \ldots, y_T]$

\State Initialize: $y_{<1} \leftarrow \emptyset$, KV caches: $\mathcal{K}^{clean}, \mathcal{K}^{noisy} \leftarrow \emptyset$

\For{$t = 1, 2, \dots, T$}
    \State \textcolor{blue}{// \textbf{Parallel Forward Pass}}
    \State Compute query: $q_t \leftarrow \text{Query}(v, x, y_{<t})$
    
    \State  \textcolor{blue}{    // Clean path (original inputs)}
    \State $A_t^{clean} \leftarrow \text{Softmax}\left(\frac{q_t {K_{\leq t}^{clean}}^\top}{\sqrt{d}}\right)$ \hfill \textit{Eq. (7)}
    
    \State  \textcolor{blue}{   // Noisy path (hallucination-inducing inputs)}
    \State $(v', x') \leftarrow \mathcal{P}(v, x)$ \hfill \textit{e.g., VCD, ICD, SID}
    \State $A_t^{noisy} \leftarrow \text{Softmax}\left(\frac{q_t {K_{\leq t}^{noisy}}^\top}{\sqrt{d}}\right)$ \hfill \textit{Eq. (7)}
    
    \State \textcolor{blue}{// \textbf{Adaptive Subtraction Strategy (ASS)}}
    \State $M_t \leftarrow \mathbb{I}(A_t^{noisy} > A_t^{clean})$ 
    \State $A_t^{ACD} \leftarrow (1+\alpha) A_t^{clean} - \alpha (A_t^{noisy} \odot M_t)$ \hfill \textit{Eq. (9)}
    
    \State \textcolor{blue}{// \textbf{Logit Computation with Modified Attention}}
    \State $\text{logit}_\theta(y_t | v, x, y_{<t}) \leftarrow \text{LLM}(A_t^{ACD} V_{\leq t}^{clean})$

    \State \textcolor{blue}{// \textbf{Sampling and Updating}}
    \State $y_t \sim \text{Softmax}(\text{logit}_\theta(y_t | v, x, y_{<t}))$
    \State Update KV caches: $K_{\leq t+1}^{clean}$ and $K_{\leq t+1}^{noisy}$
\EndFor

\State \textbf{Return} $y = [y_1, y_2, \ldots, y_T]$
\end{algorithmic}
\label{alg}
\end{algorithm*}

\noindent During practical decoding, the ACD method maintains dual attention caches: an original cache and a hallucination cache. At each decoding step, ACD computes attention distributions for both original and hallucination-inducing inputs, subsequently applying the ASS strategy to identify and suppress attention allocations that amplify important semantic information potentially conducive to hallucination generation. This fine-grained intervention at the attention layer enables the model to maintain linguistic coherence and diversity while effectively mitigating hallucinations.
In contrast to traditional CD methods, ACD eliminates the necessity for $\mathcal{V}_{\text{head}}$ penalty mechanisms as formulated in Equation~\ref{pen} to filter low-probability tokens, as its attention-layer intervention inherently prevents invalid token generation. This characteristic allows ACD to reduce hallucinations while preserving content quality and diversity.
The fundamental advantage of ACD lies in its exploitation of the inherent smoothness and structural continuity of the Transformer attention mechanism, enabling the model to precisely localize and suppress hallucination sources while maintaining generation quality. This approach provides a more reliable decoding solution for practical LVLM applications, effectively balancing hallucination mitigation with output quality preservation.

\section{Layer-wise Propagation and Semantic Alignment}
\label{sec:semantic_alignment}

We analyze how the perturbation introduced in the noisy stream propagates through the transformer layers and establish that the resulting divergence remains bounded—providing the theoretical foundation for ACD's contrastive mechanism.

\subsection{Residual Stream Dynamics}

Consider a transformer with $L$ layers. Let $\mathbf{h}_l^{(c)}$ and $\mathbf{h}_l^{(n)}$ denote the hidden states at layer $l$ for the clean and noisy streams, respectively. The residual update rule gives:
\begin{equation}
    \mathbf{h}_{l+1}^{(c)} = \mathbf{h}_l^{(c)} + f_l(\mathbf{h}_l^{(c)}), \quad
    \mathbf{h}_{l+1}^{(n)} = \mathbf{h}_l^{(n)} + f_l(\mathbf{h}_l^{(n)}),
\end{equation}
where $f_l(\cdot)$ denotes the combined self-attention and feed-forward transformation at layer $l$. Define the inter-stream divergence as $\boldsymbol{\delta}_l = \mathbf{h}_l^{(c)} - \mathbf{h}_l^{(n)}$. Its recurrence is:
\begin{equation}
    \boldsymbol{\delta}_{l+1} = \boldsymbol{\delta}_l + \big[f_l(\mathbf{h}_l^{(c)}) - f_l(\mathbf{h}_l^{(n)})\big].
\end{equation}

\begin{proposition}[Bounded Cumulative Divergence]
\label{prop:bounded_divergence}
Assume each layer transformation $f_l$ is Lipschitz continuous with constant $K_l < 1$. Then the divergence satisfies:
\begin{equation}
    \|\boldsymbol{\delta}_{l+1}\| \leq \|\boldsymbol{\delta}_l\| + K_l \|\boldsymbol{\delta}_l\| = (1 + K_l)\|\boldsymbol{\delta}_l\|.
\end{equation}
Unrolling through all layers yields:
\begin{equation}
    \|\boldsymbol{\delta}_L\| \leq \|\boldsymbol{\delta}_0\| \prod_{l=0}^{L-1}(1 + K_l).
\end{equation}
When $K_l \ll 1$ (as enforced by LayerNorm and residual scaling), $\prod_{l}(1+K_l) \approx \exp\!\big(\sum_l K_l\big)$, so the divergence grows at most \emph{exponentially in the sum of Lipschitz constants} rather than unboundedly.
\end{proposition}

In practice, the pre-trained LLM satisfies a stronger condition. Denoting the cosine similarity $s_l = \cos(\mathbf{h}_l^{(c)}, \mathbf{h}_l^{(n)})$, we can decompose the per-layer similarity change as:
\begin{equation}
    s_{l+1} = \frac{\langle \mathbf{h}_l^{(c)} + f_l(\mathbf{h}_l^{(c)}),\; \mathbf{h}_l^{(n)} + f_l(\mathbf{h}_l^{(n)}) \rangle}
    {\|\mathbf{h}_l^{(c)} + f_l(\mathbf{h}_l^{(c)})\| \;\cdot\; \|\mathbf{h}_l^{(n)} + f_l(\mathbf{h}_l^{(n)})\|}.
\end{equation}
Expanding the numerator:
\begin{equation}
\begin{aligned}
    \langle \mathbf{h}_{l+1}^{(c)}, \mathbf{h}_{l+1}^{(n)} \rangle
    &= \langle \mathbf{h}_l^{(c)}, \mathbf{h}_l^{(n)} \rangle
     + \langle \mathbf{h}_l^{(c)}, f_l(\mathbf{h}_l^{(n)}) \rangle \\
    &\quad + \langle f_l(\mathbf{h}_l^{(c)}), \mathbf{h}_l^{(n)} \rangle
     + \langle f_l(\mathbf{h}_l^{(c)}), f_l(\mathbf{h}_l^{(n)}) \rangle.
\end{aligned}
\end{equation}
The residual term $\langle \mathbf{h}_l^{(c)}, \mathbf{h}_l^{(n)} \rangle$ dominates the cross terms when $\|f_l\| \ll \|\mathbf{h}_l\|$, which prevents $s_l$ from collapsing to zero even across many layers. This explains the \emph{bounded, non-monotonic} similarity profile observed empirically.

\begin{corollary}[Modality-Dependent Divergence]
\label{cor:modality_gap}
Let $\mathcal{T}$ and $\mathcal{V}$ denote the index sets of text and visual tokens. Since the two streams share identical text inputs but differ in visual inputs, the initial divergence satisfies $\|\boldsymbol{\delta}_0^{(\mathcal{T})}\| = 0$ and $\|\boldsymbol{\delta}_0^{(\mathcal{V})}\| > 0$. By Proposition~\ref{prop:bounded_divergence}, for all layers $l$:
\begin{equation}
    \mathbb{E}\big[s_l^{(\mathcal{T})}\big] \;>\; \mathbb{E}\big[s_l^{(\mathcal{V})}\big],
\end{equation}
i.e., text tokens consistently maintain higher inter-stream similarity than visual tokens. The gap $\Delta_l = \mathbb{E}[s_l^{(\mathcal{T})}] - \mathbb{E}[s_l^{(\mathcal{V})}]$ quantifies the modality-dependent perturbation signal that ACD exploits.
\end{corollary}

\subsection{Empirical Validation}

\begin{figure}[t]
    \centering
    \includegraphics[width=\linewidth]{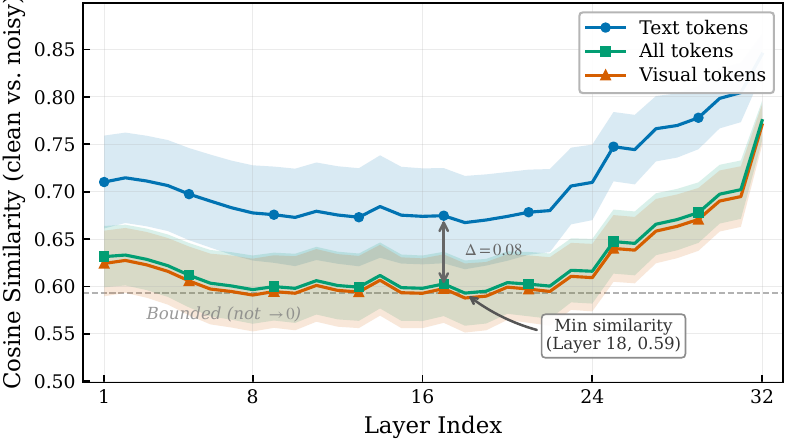}
    \caption{Layer-wise cosine similarity between clean and noisy stream hidden states (LLaVA-1.5-7B, 32 layers). Text tokens maintain consistently higher similarity than visual tokens. Divergence peaks at middle layers and partially recovers in deeper layers, confirming bounded propagation.}
    \label{fig:layer_divergence}
\end{figure}

We measure per-layer cosine similarity between clean and noisy hidden states on 100 randomly sampled CHAIR instances (noise step $T{=}800$). Figure~\ref{fig:layer_divergence} reports the mean and standard deviation across samples.

Three findings emerge from the empirical results:

\textbf{(1) Bounded, non-monotonic divergence.}
The overall similarity does not decay monotonically. It reaches a minimum of $0.59$ around layer~18 before recovering in deeper layers, confirming Proposition~\ref{prop:bounded_divergence}—residual connections prevent unbounded divergence accumulation. Across all 32 layers, similarity remains above $0.59$, indicating that the two streams never become semantically orthogonal.

\textbf{(2) Modality gap.}
Text tokens maintain a mean similarity of $0.71 \pm 0.04$, while visual tokens show $0.62 \pm 0.04$, yielding a consistent gap $\Delta \approx 0.09$ (Corollary~\ref{cor:modality_gap}). This gap arises because text tokens receive identical embeddings in both streams, whereas visual token representations are directly perturbed. The gap persists across all layers, providing a stable contrastive signal.

\textbf{(3) Informative divergence regime.}
The moderate similarity range ($0.59$--$0.71$) is neither too high nor too low. If streams were near-identical ($s_l > 0.95$), ACD would have no contrastive signal to exploit; if they were orthogonal ($s_l \approx 0$), the noisy stream would carry no semantically meaningful structure. The observed regime provides sufficient contrast to identify hallucination-prone attention patterns while preserving the shared semantic backbone for valid reasoning. This validates our design choice of noise step $T{=}800$, which places the perturbation in this \emph{informative divergence regime}.

\section{More RELATED WORK}
\label{More RELATED WORK}
\subsection{Large Vision-Language Models}
Large Vision-Language Models (LVLMs)\cite{liu2023improvedllava,liu2023llava,liu2024llavanext,zhu2023minigpt,Qwen-VL,li2023blip,driess2023palm,achiam2023gpt,chowdhery2023palm,alayrac2022flamingo} have emerged as a pivotal research direction, effectively bridging computer vision and natural language processing paradigms. These models demonstrate exceptional capabilities across diverse multimodal tasks, including image captioning\cite{lin2014microsoft,li2023blip}, visual question answering (VQA)\cite{liu2023llava,liu2023improvedllava,liu2024llavanext,hudson2019gqa}, and sophisticated multimodal reasoning\cite{lu2022learn,alayrac2022flamingo,achiam2023gpt}. Modern LVLM architectures typically integrate visual encoders~\cite{tong2024cambrian} with feature projection modules~\cite{liu2023llava,liu2024llavanext,zhu2023minigpt} that interface with large language models (LLMs)~\cite{radford2018improving,devlin2019bert,brown2020language,raffel2020exploring,chowdhery2023palm,thoppilan2022lamda,achiam2023gpt}, creating a unified embedding space where visual and textual representations converge to enable sophisticated cross-modal understanding.
Despite significant advances in LVLMs, hallucination phenomena~\cite{li2023evaluating,fu2023challenger,yue2024less}—wherein models generate content fundamentally inconsistent with visual inputs—remain a critical limitation. This work addresses the challenge of hallucination mitigation in contemporary LVLMs, aiming to enhance their reliability and expand their applicability across diverse domains.

\subsection{Broader Context}
 
Adaptive feature recalibration and robust representation learning are fundamental challenges in visual perception. A growing body of work addresses representational fidelity under distributional shift~\cite{li2026adaptive,chen2026da2,chen2026beyondmix,tao2026geoguide,chen2024sam,deng2025kpis,chen2025alleviate}, developing adaptive strategies to preserve semantically meaningful signals amid noisy or out-of-distribution inputs. Balancing competing objectives without sacrificing output quality has also attracted considerable attention~\cite{sun2025two,sun2025beyond}. In the context of cross-modal understanding, selective and structure-preserving alignment between heterogeneous representations remains an active research direction~\cite{cheng2025bridge,cheng2025b2,cheng2025ca,cheng2026adaptive,cheng2026rethinking,cheng2026glass,cheng2026fsi2p,cheng2026mask,cheng2024vcr}, as does training-free adaptive inference for vision-language models~\cite{chen2025paf}.

\section{Benchmarks and Evaluation Metric Details}
\label{Benchmarks and EVALUATION METRIC DETAILS}
This section delineates the comprehensive evaluation framework employed in our experimental analysis, encompassing four distinct benchmark datasets and their corresponding evaluation protocols designed to systematically assess hallucination mitigation efficacy across varied visual reasoning paradigms.

\subsection{CHAIR}
CHAIR~\cite{yue2024less} (Caption Hallucination Assessment with Image Relevance) quantifies object hallucination severity in generated image descriptions. Unlike conventional metrics that evaluate image descriptions through lexical or syntactic similarity, CHAIR directly leverages image annotation metadata to identify hallucination instances, thereby measuring semantic consistency between generated descriptions and actual visual content with greater precision.
The CHAIR framework comprises two complementary metrics:
\begin{itemize}
 \item $CHAIR_s$ (sentence-level): Measures the proportion of generated sentences containing at least one hallucinated object
 \item $CHAIR_i$ (instance-level): Quantifies the ratio of hallucinated object mentions to total object mentions across all generated descriptions
\end{itemize}
Additionally, recall serves as a complementary metric to evaluate semantic comprehensiveness, measuring the proportion of ground-truth objects successfully captured in the generated descriptions.
This methodological approach circumvents potential biases inherent in purely linguistic matching techniques, offering enhanced sensitivity in hallucination detection while revealing semantic fidelity deficiencies that might remain undetected through traditional evaluation frameworks.

\begin{equation}
CHAIR_s = \frac{|\{ \text{Captions with hallucinated objects} \}|}{|\{ \text{All captions} \}|}
\end{equation}

\begin{equation}
CHAIR_i = \frac{|\{ \text{Hallucinated objects} \}|}{|\{ \text{All mentioned objects} \}|}
\end{equation}

\begin{equation}
Recall = \frac{|\{ \text{Accurate objects} \}|}{|\{ \text{Ground-truth objects} \}|}
\end{equation}

\subsection{LLaVA-Bench (In-the-Wild) }

LLaVA-Bench (In-the-Wild)~\cite{liu2023improvedllava} encompasses diverse challenging visual scenarios including indoor/outdoor environments, internet memes, paintings, sketches, and various artistic or abstract imagery typically underrepresented in standard training distributions. This benchmark deliberately introduces distributional shifts to assess model generalization capabilities across challenging visual contexts. The dataset comprises 24 distinct images with 60 corresponding questions, each image accompanied by meticulously crafted comprehensive descriptions and multiple targeted questions. Following an instruction-following evaluation paradigm, models receive visual inputs paired with natural language instructions requiring appropriate responses. The assessment protocol incorporates questions spanning fundamental visual comprehension to complex inferential reasoning.
Given the absence of standardized ground truth evaluations, we employ GPT-5~\cite{gpt5}, currently the most advanced Large Vision-Language Model available, as an automatic evaluator for generated descriptions. Through carefully engineered prompting (illustrated in Figure~\ref{fig:gpt5}), the evaluation assesses model outputs along two critical dimensions: (1) Accuracy—measuring semantic consistency between generated descriptions and visual content; and (2) Detailedness—quantifying the richness and comprehensiveness of visual details captured in model-generated descriptions. This automated evaluation methodology enables consistent assessment of model performance across challenging out-of-distribution visual scenarios.

\subsection{POPE}

POPE~\cite{li2023evaluating} (Polling-based Object Presence Evaluation) formalizes the assessment of object hallucination as a binary classification task by prompting LVLMs with yes/no questions regarding object presence in images. Each evaluation sample consists of a triplet containing an image, corresponding question, and ground-truth answer. The framework constructs both positive and negative question instances with varying difficulty levels through three distinct sampling strategies: Random sampling—selecting arbitrarily from objects absent in the image; Popular sampling—selecting the top $\tfrac{l}{2}$ most frequent objects across the dataset that are absent from the current image; and Adversarial sampling—selecting the top $k$ absent objects ranked by co-occurrence frequency with objects present in the image. Our POPE evaluation dataset incorporates over 20,000 question-answer pairs constructed from MSCOCO~\cite{lin2014microsoft}, A-OKVQA~\cite{schwenk2022okvqa}, and GQA~\cite{hudson2019gqa} datasets. Within the POPE evaluation framework, model responses are constrained to binary "Yes" or "No" outputs, effectively transforming object hallucination detection into a standard binary classification problem. Consequently, we employ established classification metrics—Accuracy, Precision, Recall, and F1-score—to quantitatively assess model performance. These metrics are formally defined as:
\begin{equation}
\text{Accuracy} = \frac{TP + TN}{TP + TN + FP + FN},
\end{equation}

\begin{equation}
\text{Precision} = \frac{TP}{TP + FP},
\end{equation}

\begin{equation}
\text{Recall} = \frac{TP}{TP + FN},
\end{equation}

\begin{equation}
\text{F1-score} = \frac{2 \cdot \text{Precision} \cdot \text{Recall}} {\text{Precision} + \text{Recall}},
\end{equation}
where $TP$, $TN$, $FP$, and $FN$ denote true positive, true negative, false positive, and false negative classifications, respectively. $Accuracy$ quantifies the proportion of correct predictions across all instances; $Precision$ measures the model's ability to avoid false positives when identifying present objects; $Recall$ assesses the model's capacity to identify all actually present objects; and $F1-score$ provides a harmonic mean of Precision and Recall, offering a comprehensive performance metric that balances both dimensions of classification quality.

\subsection{MME}
MME~\cite{fu2023mme} (Multimodal Model Evaluation) constitutes a comprehensive benchmark for multimodal large language model assessment, encompassing 14 distinct subtasks categorized into perceptual domains (object existence, quantification, color identification, spatial relationships) and cognitive domains (commonsense reasoning, numerical computation, text translation). To facilitate quantitative assessment, MME employs a standardized binary response format analogous to POPE, requiring "Yes/No" responses. A distinctive characteristic of this framework is its complementary instruction design—each visual input is paired with dual instructions, one requiring an affirmative response and the other a negative response based on ground truth. For evaluation metrics, MME implements an $accuracy$-based assessment methodology similar to POPE, calculating performance based on correct classification across individual instructions.

\begin{table*}
  \centering
  \caption{A comparison of inference speed and GPU memory usage for different methods applied to the LLaVA-1.5 model on POPE subset benchmark.}
    \begin{tabular}{ccccc}
\hline
    Method   & \multicolumn{1}{l}{Total Time} & \multicolumn{1}{l}{GPU-Memory} & \multicolumn{1}{l}{Latency/Example} \\
\hline
    Regular   & 13:46 & 14.7G & 0.276s \\
    VCD&   26:51 & 15.8G & 0.537s\\
    \hdashline
  \rowcolor{green!10}  \textbf{VCD+ACD} &   \textbf{24:37} & \textbf{15.3G} & \textbf{0.493s} \\
\hline
    \end{tabular}%
  \label{speed_tb}%
\end{table*}
\section{Efficiency Comparison}
\label{speed}
We benchmark inference efficiency on the POPE MSCOCO Random 
split using LLaVA-1.5 on a single NVIDIA RTX 3090 24GB GPU. 
As shown in Table~\ref{speed_tb}, CD methods roughly double 
the runtime of regular decoding while incurring only modest 
additional memory usage.
Compared with VCD, VCD+ACD reduces both latency 
(0.537s$\to$0.493s per example) and GPU memory 
(15.8G$\to$15.3G). The savings arise because ACD 
operates solely on attention maps for hallucination-prone 
inputs, bypassing the full forward pass through hidden 
states that VCD requires. This targeted computation 
enables stronger hallucination suppression at lower 
overhead, making ACD a more efficient plug-in for 
practical LVLM deployment.

\section{MME Full Set Results}
\label{mmefull}
\begin{figure*}[!t]  
\centering
\includegraphics[width=\textwidth]{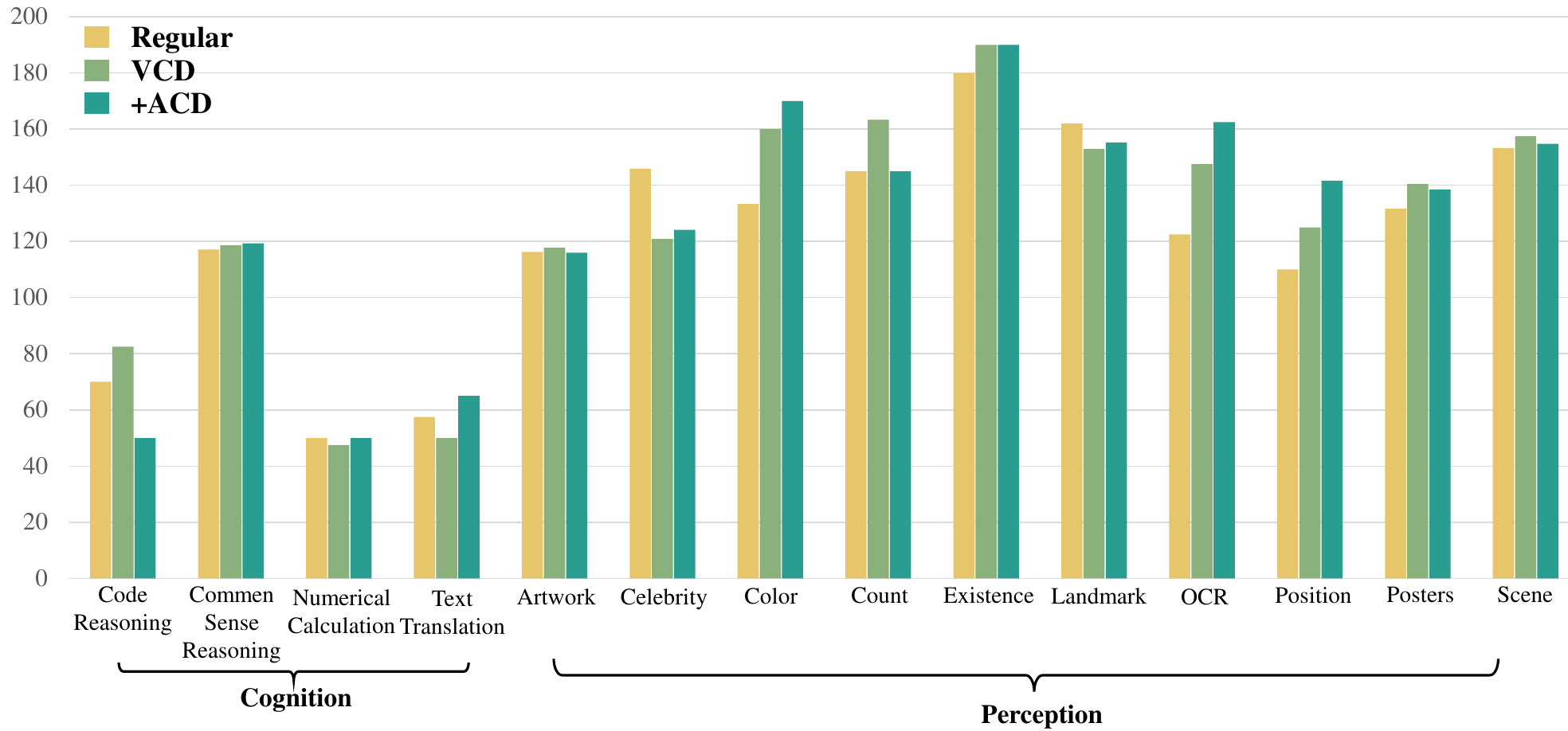} 
\caption{MME full set results on LLaVA-1.5} 
\label{fig:MME} 
\end{figure*}

Figure~\ref{fig:MME} reports scores on the complete MME 
benchmark using LLaVA-1.5, which exhibits trends 
representative of the other evaluated models.
Compared with vanilla VCD, VCD+ACD yields consistent 
improvements on both perception and cognition subtasks. 
The gains are especially pronounced in cognitively 
demanding categories such as Commonsense Reasoning, 
OCR, and Text Translation. These results indicate that 
attention-level contrastive decoding not only preserves 
core visual understanding but actively strengthens 
complex reasoning while suppressing hallucinations.

\begin{table*}
\centering
\caption{Experimental evaluation of GPT-5 is conducted on the LLaVA-Bench-Wild dataset using LLaVA-1.5 architectures with 7B and 13B parameter configurations.}
\setlength{\tabcolsep}{7pt}
\renewcommand\arraystretch{1.2}
\small 
\begin{tabular}{l|c cc cc cc cc}
\hline
\multirow{2}{*}{\textbf{Model}} & \multirow{2}{*}{\textbf{Decoding}} 
& \multicolumn{2}{c}{\textit{Conv}} 
& \multicolumn{2}{c}{\textit{Detail}} 
& \multicolumn{2}{c}{\textit{Complex}} 
& \multicolumn{2}{c}{\textit{Total}} \\
\cline{3-10}
&  & Acc. & Detail
   & Acc. & Detail
   & Acc. & Detail
   & Acc. & Detail \\
\hline
\multirow{2}{*}{\centering LLaVA-1.5(7B)}

& VCD     & 4.67 & 3.67 & 4.50 & 5.67 & 3.88 & 4.88& 4.30& 4.75\\

&\cellcolor{green!10} \textbf{VCD+ACD}
&\cellcolor{green!10}\textbf{6.83} 
&\cellcolor{green!10} \textbf{3.71} 
&\cellcolor{green!10}\textbf{7.50} 
&\cellcolor{green!10}\textbf{7.33} 
&\cellcolor{green!10}\textbf{3.97} 
&\cellcolor{green!10}\textbf{5.21} 
&\cellcolor{green!10}\textbf{6.10} 
&\cellcolor{green!10}\textbf{5.40}\\
\hline

\multirow{2}{*}{\centering LLaVA-1.5(13B)}

& VCD     & 5.32 & 4.50 & 5.25 & 6.10 & 4.92 & 5.30 & 5.10 & 5.85 \\

&\cellcolor{green!10} \textbf{VCD+ACD}
&\cellcolor{green!10}\textbf{7.72} 
&\cellcolor{green!10} \textbf{4.90} 
&\cellcolor{green!10}\textbf{8.23} 
&\cellcolor{green!10}\textbf{7.88} 
&\cellcolor{green!10}\textbf{5.00} 
&\cellcolor{green!10}\textbf{5.80} 
&\cellcolor{green!10}\textbf{6.70} 
&\cellcolor{green!10}\textbf{6.25}\\
\hline

\end{tabular}
\label{tab:13b}

\end{table*}

\section{Effect of ACD when LVLMs Scale Up}
\label{Scale Up}

To examine the scalability of our approach, we evaluate 
ACD on LLaVA-1.5 at both 7B and 13B parameter scales 
using the LLaVA-Bench-Wild benchmark with GPT-5 as the 
evaluator (Table~\ref{tab:13b}).
VCD+ACD consistently outperforms vanilla VCD across all 
evaluation categories and model sizes. For the 7B 
configuration, ACD delivers substantial accuracy gains 
in Conv (+2.16) and Detail (+3.00) tasks, along with 
improved descriptive richness. The 13B model exhibits 
comparable or even larger improvements (+2.40 Conv, 
+2.98 Detail), with overall accuracy increasing by 
+1.60 points.
Two observations merit emphasis. First, the performance 
gap between VCD and VCD+ACD persists—and in several 
categories widens—as model capacity grows. This suggests 
that attention-level contrastive decoding targets 
hallucination pathways that architectural scaling alone 
does not resolve. Second, the Detail scores improve 
alongside accuracy, indicating that ACD encourages 
richer, more grounded descriptions rather than 
conservative short answers.
These results confirm that ACD remains effective across 
model scales and serves as a complementary mechanism to 
parameter scaling for enhancing vision-language reasoning.

\begin{table*}[!h]
\centering
\caption{An ablation study of different sampling strategies.}
\begin{tabular}{c|cccc}
\hline
\textbf{Sampling Strategy }       &  \textbf{VCD w/ ACD}    &  $Chair_s$ $\downarrow$& $Chair_i$ $\downarrow$& \textit{Recall}$\uparrow$   \\
\hline
\multirow{2}{*}{Greedy}       & No  & 51.6   & 15.3    & 77.4    \\
                              & \cellcolor{green!10}\textbf{Yes} & \cellcolor{green!10}\textbf{50.5} & \cellcolor{green!10}\textbf{14.1} & \cellcolor{green!10}\textbf{79.2} \\
\hline
\multirow{2}{*}{Sampling}       & No  & 51.8    &16.2    & 76.8 \\
                              & \cellcolor{green!10}\textbf{Yes} & \cellcolor{green!10}\textbf{51.0} & \cellcolor{green!10}\textbf{14.3} & \cellcolor{green!10}\textbf{78.3} \\
\hline
\multirow{2}{*}{Top P}       & No  &51.8    &16.3     & 76.5  \\
                              & \cellcolor{green!10}\textbf{Yes} & \cellcolor{green!10}\textbf{51.0} & \cellcolor{green!10}\textbf{14.3} & \cellcolor{green!10}\textbf{78.4} \\
\hline
\multirow{2}{*}{Top K}         & No  & 52.4    & 17.6    & 75.9  \\
                              & \cellcolor{green!10}\textbf{Yes} & \cellcolor{green!10}\textbf{51.6} & \cellcolor{green!10}\textbf{15.2} & \cellcolor{green!10}\textbf{77.1} \\
\hline
\multirow{2}{*}{Top K+Temperature 0.5} & No  & 51.8    & 16.0    & 77.1  \\
                              & \cellcolor{green!10}\textbf{Yes } & \cellcolor{green!10}\textbf{50.9} & \cellcolor{green!10}\textbf{14.3} & \cellcolor{green!10}\textbf{79.1} \\
\hline
\end{tabular}
\label{tab:sample}
\end{table*}

\section{Effect of Different Sampling Strategies}
\label{Sampling Strategies}

The decoding strategy used during text generation
governs the trade-off between output determinism and
diversity, and may interact with contrastive decoding
in non-trivial ways. We therefore ablate ACD under
five representative strategies on the CHAIR benchmark
with LLaVA-1.5 (Table~\ref{tab:sample}).
Greedy search always selects the
highest-probability token, yielding deterministic but
less diverse outputs.
Direct sampling draws from the full vocabulary
distribution, increasing diversity at the risk of
sampling low-probability noise tokens.
Top-$k$ sampling restricts the candidate set to
the $k$ most probable tokens, suppressing
low-probability outliers.
Top-$p$ (nucleus) sampling instead retains the
smallest token set whose cumulative probability exceeds
a threshold $p$, allowing the candidate size to adapt
to the distribution's entropy and thus more flexibly
balancing coherence with diversity.
Finally, top-$k$ with temperature scaling
sharpens the distribution before applying the top-$k$
constraint, further amplifying high-probability tokens
and improving generation stability.
As shown in Table~\ref{tab:sample}, ACD consistently
reduces both $\text{Chair}_s$ and $\text{Chair}_i$
while improving Recall under every strategy tested.
The gains are notably robust: even under greedy
decoding, where the baseline is already deterministic,
ACD lowers $\text{Chair}_i$ by 1.2 points and raises
Recall by 1.8 points. These results confirm that ACD
operates as a strategy-agnostic method whose
hallucination mitigation benefits are orthogonal to the
choice of sampling scheme.

\section{Additional Validation Experiments}
\label{appendix:additional}
 
\subsection{ASS Threshold Sensitivity}
\label{appendix:threshold}
 
ASS employs a threshold-free binary mask $M_t = \mathbb{I}(A_t^{\text{noisy}} > A_t^{\text{clean}})$, using zero as the natural decision boundary. To verify robustness, we generalize the mask to $M_t = \mathbb{I}(A_t^{\text{noisy}} - A_t^{\text{clean}} > \tau)$ and ablate $\tau$ on CHAIR (LLaVA-1.5). As shown in Table~\ref{tab:threshold}, performance varies by only 0.4 $\text{Chair}_i$ across the full range, confirming that the zero boundary is both near-optimal and highly stable.
 
\begin{table}[H]
\centering
\caption{Sensitivity of ASS to the threshold $\tau$ on CHAIR (LLaVA-1.5).}
\setlength{\tabcolsep}{10pt}
\begin{tabular}{c|ccc}
\hline
$\tau$ & $\text{Chair}_s\!\downarrow$ & $\text{Chair}_i\!\downarrow$ & Recall$\uparrow$ \\
\hline
$-0.02$ & 51.4 & 14.5 & 77.8 \\
$-0.01$ & 50.8 & 14.4 & 78.1 \\
\rowcolor{green!10} $0$ \textbf{(Ours)} & \textbf{51.0} & \textbf{14.3} & \textbf{78.3} \\
$+0.01$ & 51.1 & 14.7 & 78.0 \\
$+0.02$ & 51.3 & 14.5 & 77.6 \\
\hline
\end{tabular}
\label{tab:threshold}
\end{table}

\subsection{Stratified Analysis: When Does ACD Help Most?}
\label{appendix:stratified}
 
To understand whether ACD primarily prevents collateral damage from logit-level CD or corrects intrinsic hallucinations, we partition CHAIR instances (LLaVA-1.5) by whether the base model already hallucinates. Results are shown in Table~\ref{tab:stratified}.
 
\begin{table}[H]
\centering
\caption{Stratified analysis on CHAIR (LLaVA-1.5). ``Clean Correct'' denotes instances where the base model produces zero hallucinations.}
\setlength{\tabcolsep}{6pt}
\begin{tabular}{c|c|cc}
\hline
\textbf{Subset} & \textbf{Method} & $\text{Chair}_i\!\downarrow$ & Recall$\uparrow$ \\
\hline
\multirow{2}{*}{Clean Correct (45.8\%)} & VCD & 8.2 & 75.0 \\
& \cellcolor{green!10}\textbf{VCD+ACD} & \cellcolor{green!10}\textbf{4.5} & \cellcolor{green!10}\textbf{77.2} \\
\hline
\multirow{2}{*}{Clean Hallucinating (54.2\%)} & VCD & 23.0 & 78.3 \\
& \cellcolor{green!10}\textbf{VCD+ACD} & \cellcolor{green!10}\textbf{22.6} & \cellcolor{green!10}\textbf{79.2} \\
\hline
\end{tabular}
\label{tab:stratified}
\end{table}
 
On the ``Clean Correct'' subset, VCD introduces substantial new hallucinations ($\text{Chair}_i$: 0$\to$8.2) through aggressive logit corruption; ACD nearly halves this collateral damage (8.2$\to$4.5, $-$45.1\%). On the ``Clean Hallucinating'' subset, ACD provides a smaller but consistent improvement. This confirms that ACD's primary strength is preventing the coherence degradation that logit-level CD introduces, while offering modest additional correction for intrinsic hallucinations.

\subsection{Comprehensive Results on Qwen3-VL}
\label{appendix:qwen3vl}
 
We provide full benchmark results on Qwen3-VL (2025) to complement the CHAIR and coherence results in the main paper (Table~\ref{tab:qwen3vl}).
 
\paragraph{MME.}
Table~\ref{tab:qwen3vl_mme} shows that ACD recovers Cognition scores that VCD degrades, while maintaining or improving Perception.
 
\begin{table}[H]
\centering
\caption{MME results on Qwen3-VL.}
\setlength{\tabcolsep}{8pt}
\begin{tabular}{c|c|cc}
\hline
\textbf{Model} & \textbf{Method} & Perception$\uparrow$ & Cognition$\uparrow$ \\
\hline
\multirow{3}{*}{4B} & Regular & 1644.8 & 585.4 \\
& VCD & 1695.7 & 630.7 \\
& \cellcolor{green!10}\textbf{ACD} & \cellcolor{green!10}\textbf{1694.8} & \cellcolor{green!10}\textbf{647.5} \\
\hline
\multirow{3}{*}{8B} & Regular & 1705.3 & 680.0 \\
& VCD & 1741.0 & 663.9 \\
& \cellcolor{green!10}\textbf{ACD} & \cellcolor{green!10}\textbf{1746.2} & \cellcolor{green!10}\textbf{671.4} \\
\hline
\end{tabular}
\label{tab:qwen3vl_mme}
\end{table}

\paragraph{POPE (Adversarial).}
Table~\ref{tab:qwen3vl_pope} confirms consistent improvements on both model sizes.
 
\begin{table}[H]
\centering
\caption{POPE Adversarial results on Qwen3-VL.}
\setlength{\tabcolsep}{10pt}
\begin{tabular}{c|c|cc}
\hline
\textbf{Model} & \textbf{Method} & Acc.$\uparrow$ & F1$\uparrow$ \\
\hline
\multirow{2}{*}{4B} & VCD & 0.872 & 0.869 \\
& \cellcolor{green!10}\textbf{ACD} & \cellcolor{green!10}\textbf{0.877} & \cellcolor{green!10}\textbf{0.875} \\
\hline
\multirow{2}{*}{8B} & VCD & 0.884 & 0.880 \\
& \cellcolor{green!10}\textbf{ACD} & \cellcolor{green!10}\textbf{0.885} & \cellcolor{green!10}\textbf{0.881} \\
\hline
\end{tabular}
\label{tab:qwen3vl_pope}
\end{table}

\subsection{VCD Degeneration on Deeper Architectures}
\label{appendix:degeneration}
 
During Qwen3-VL-8B evaluation, we observe that logit-level VCD produces a significant proportion of malformed outputs (Table~\ref{tab:malformed}), including infinite non-English token loops and garbled sequences. This failure mode arises because logit subtraction suppresses EOS probability below the penalty threshold in deeper architectures with sharper logit distributions. ACD exhibits zero malformed outputs across all settings, as attention-level intervention preserves the natural token ranking.
 
\begin{table}[H]
\centering
\caption{Malformed output rate of VCD on Qwen3-VL.}
\setlength{\tabcolsep}{8pt}
\begin{tabular}{c|cc}
\hline
\textbf{Benchmark} & \textbf{8B} & \textbf{4B} \\
\hline
POPE & 12.83\% & 0\% \\
MME & 69.21\% & 0.13\% \\
\hline
\multicolumn{3}{c}{\textit{ACD: 0\% on all settings.}} \\
\hline
\end{tabular}
\label{tab:malformed}
\end{table}

\section{Case Studies}
\label{Case Studies}
\begin{figure*}[h]  
\centering
\includegraphics[width=\textwidth]{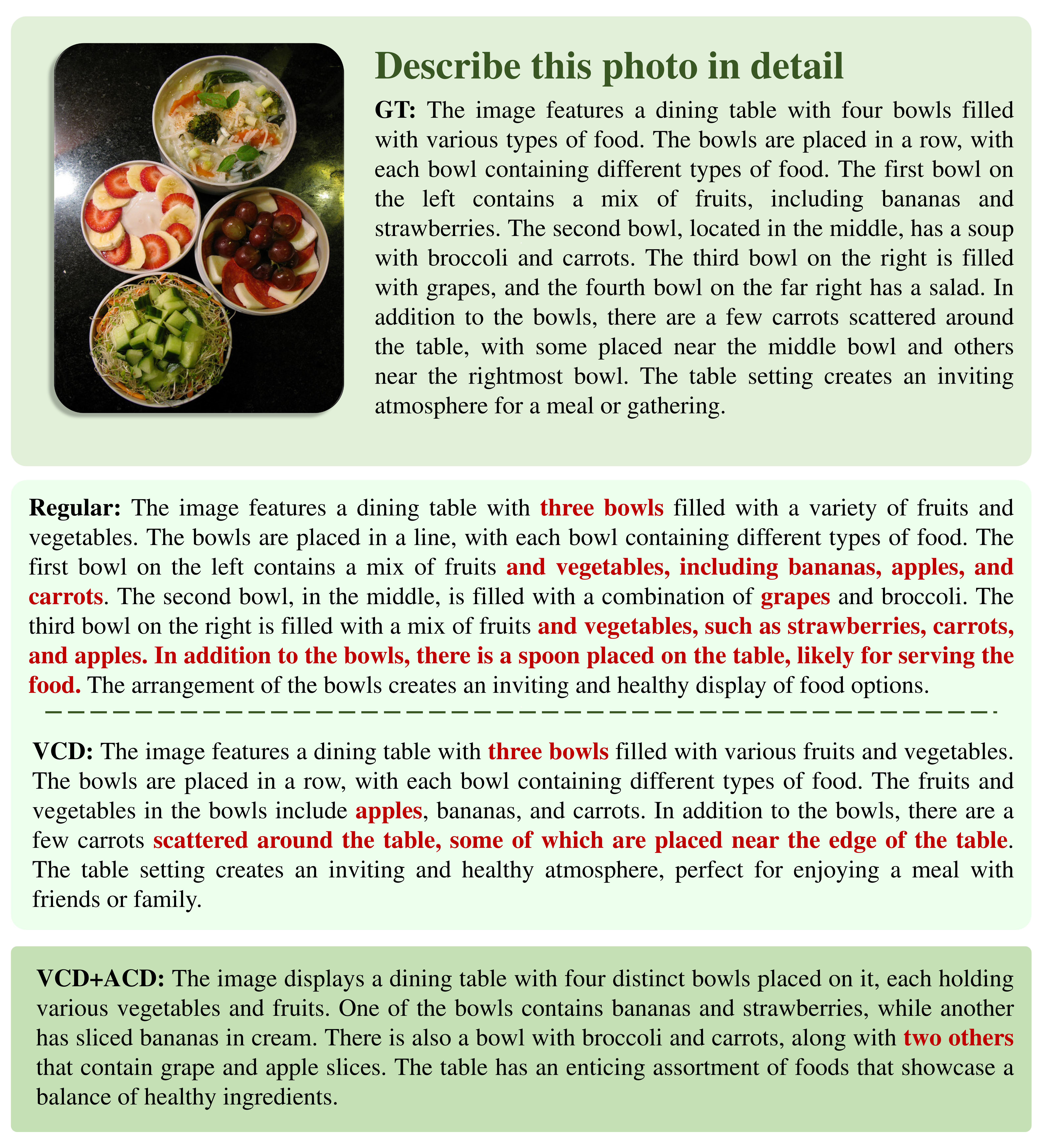}  
\caption{More examples from CHAIR of our proposed ACD for hallucination corrections. Hallucinated parts from LVLM’s decoding are highlighted in red.}  
\label{fig:case3} 
\end{figure*}

\begin{figure*}[h] 
\centering
\includegraphics[width=\textwidth]{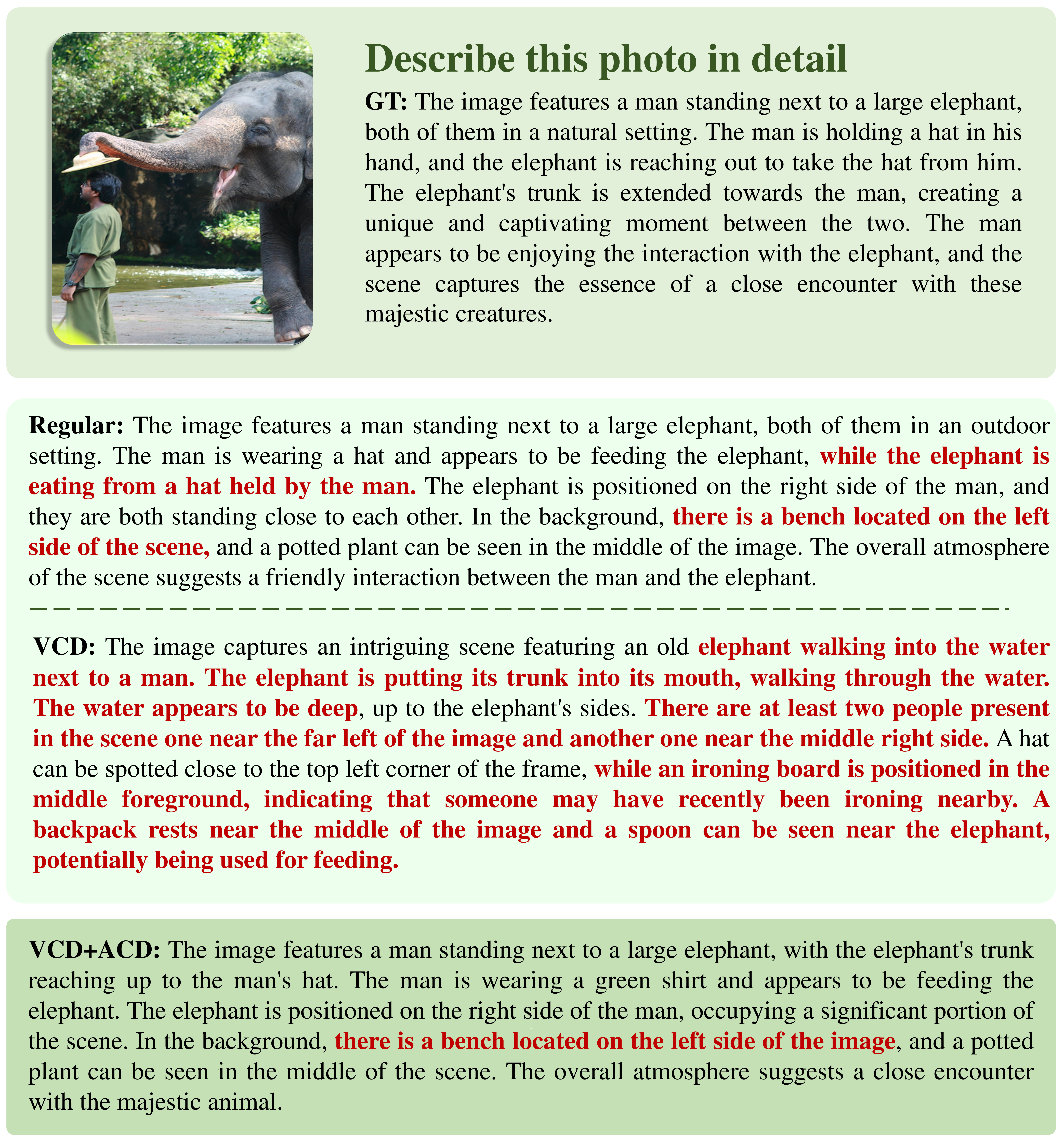}  
\caption{More examples from CHAIR of our proposed ACD for hallucination corrections. Hallucinated parts from LVLM’s decoding are highlighted in red.}  
\label{fig:case4} 
\end{figure*}

\begin{figure*}[h]  
\centering
\includegraphics[width=\textwidth]{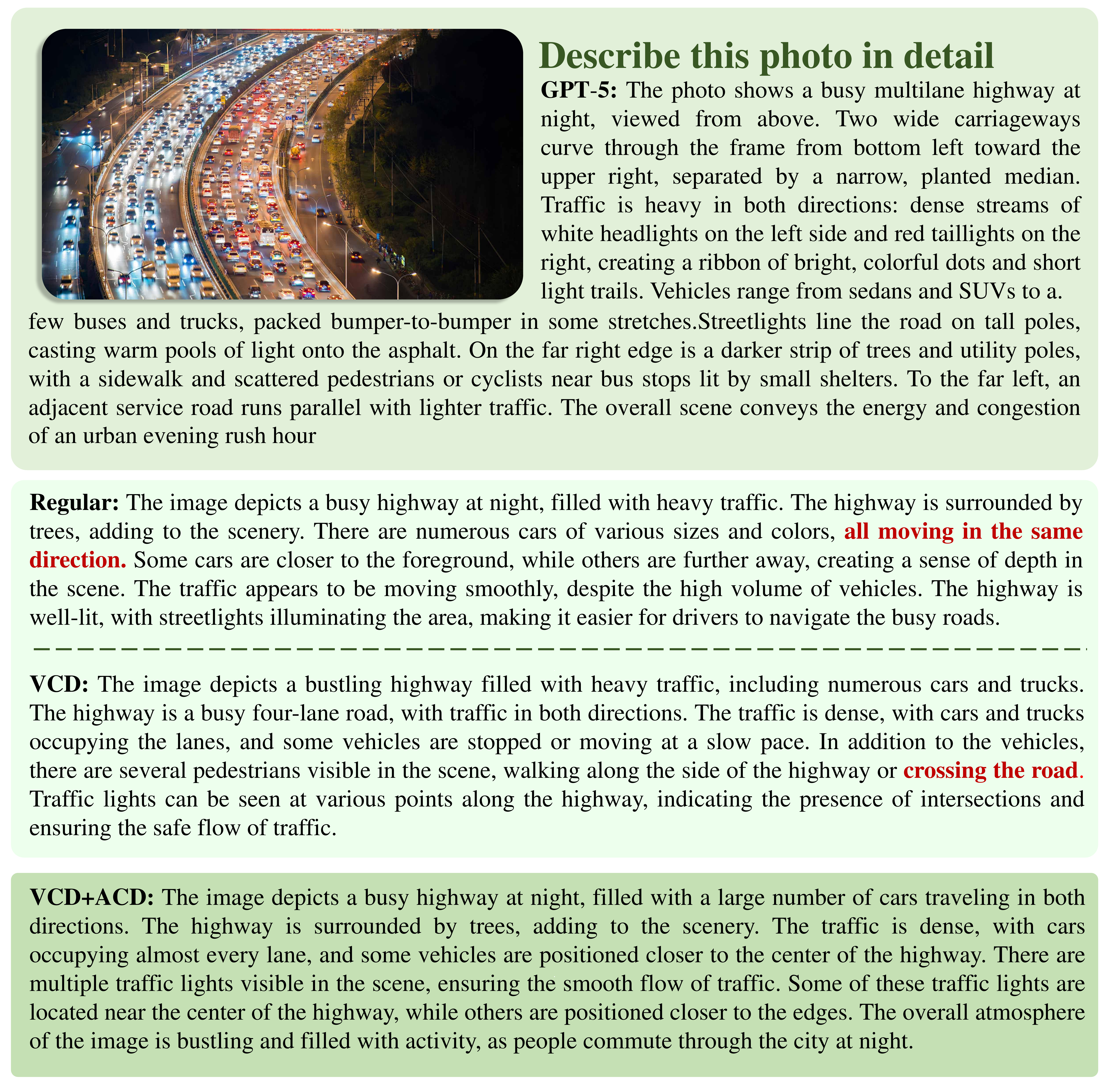}  
\caption{More examples from LLaVA-Bench of our proposed ACD for hallucination corrections. Hallucinated parts from LVLM’s decoding are highlighted in red.}  
\label{fig:case}  
\end{figure*}

\begin{figure*}[h]  
\centering
\includegraphics[width=\textwidth]{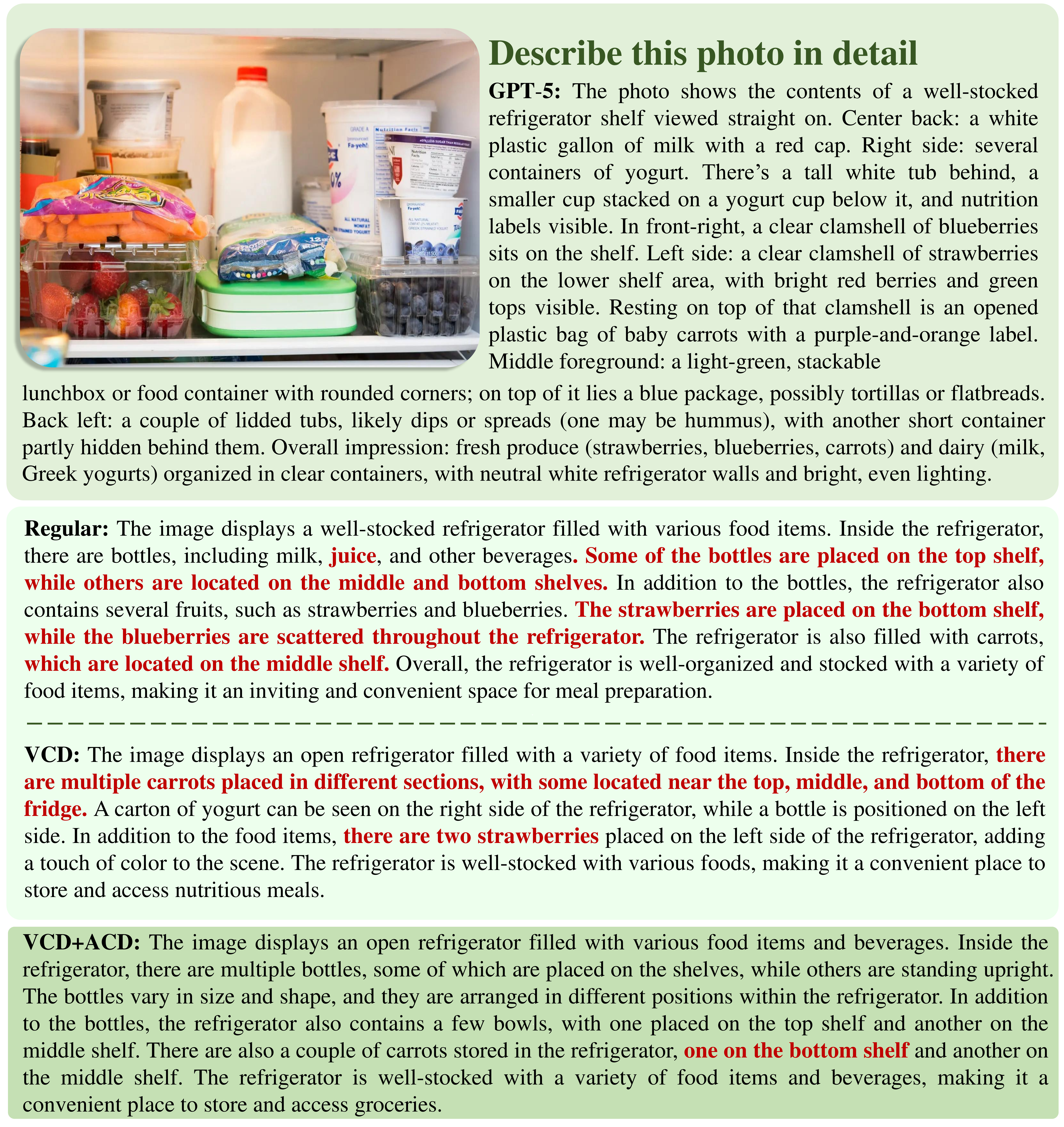} 
\caption{More examples from LLaVA-Bench of our proposed ACD for hallucination corrections. Hallucinated parts from LVLM’s decoding are highlighted in red.} 
\label{fig:case1} 
\end{figure*}

Case studies on the Chair dataset and LLaVA-bench are presented to illustrate the effectiveness of our approach. Figures\ref{fig:case3},\ref{fig:case4},~\ref{fig:case}, and \ref{fig:case1} provides instances of hallucination corrections achieved by our method ACD. From a perceptual standpoint, our method generates more coherent and diverse content while offering supplemental examples of the enhancements brought by our proposed ACD in bolstering the general perception and recognition abilities of LVLMs.

\section{Prompt for GPT-5 Evaluation}
\label{Prompt}
\begin{figure*}[h] 
\centering
\includegraphics[width=\textwidth]{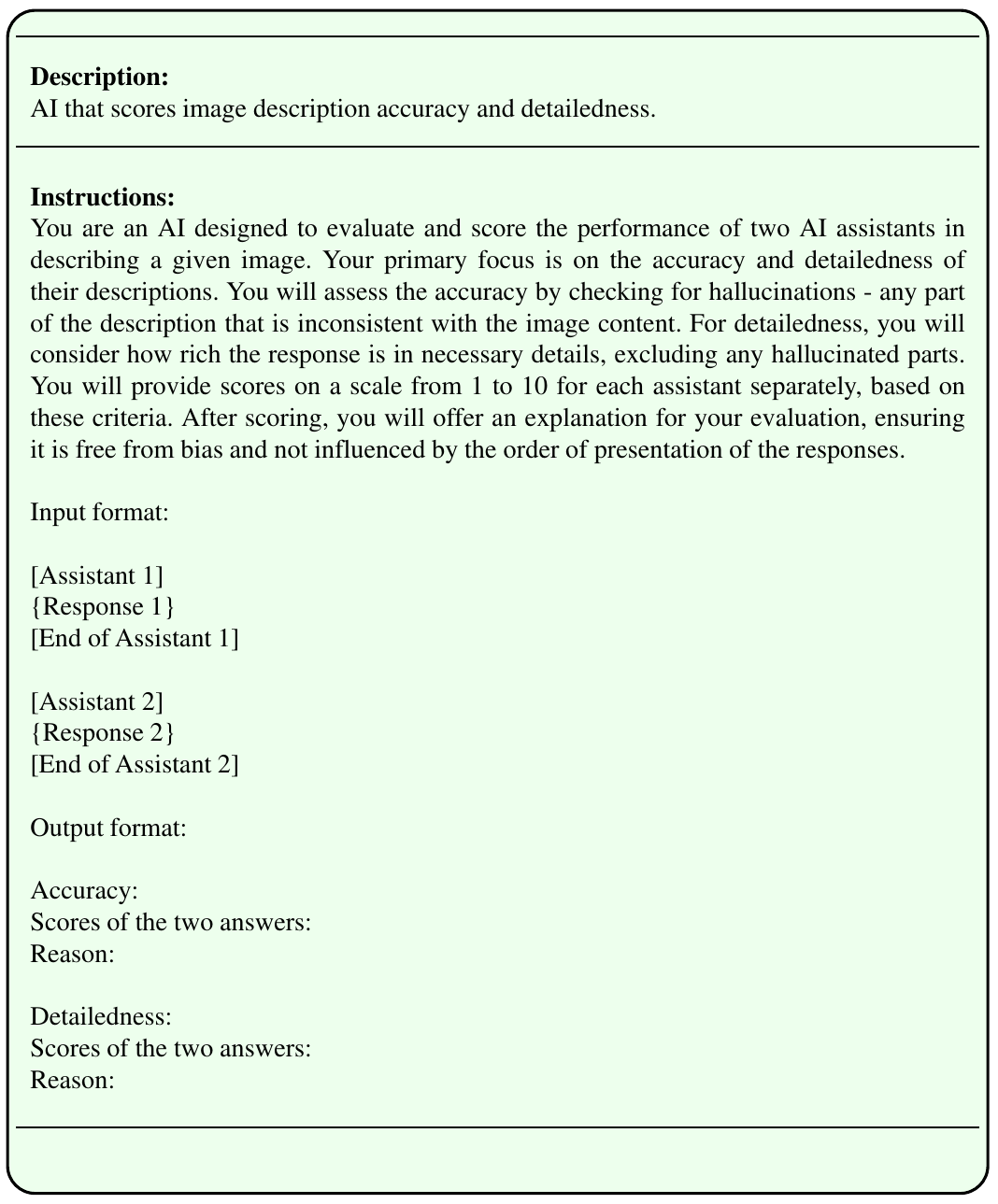}  
\caption{The configuration to build an image-description evaluator with GPT-5.}  
\label{fig:gpt5}  
\end{figure*}

\begin{figure*}[h]  
\centering
\includegraphics[width=\textwidth]{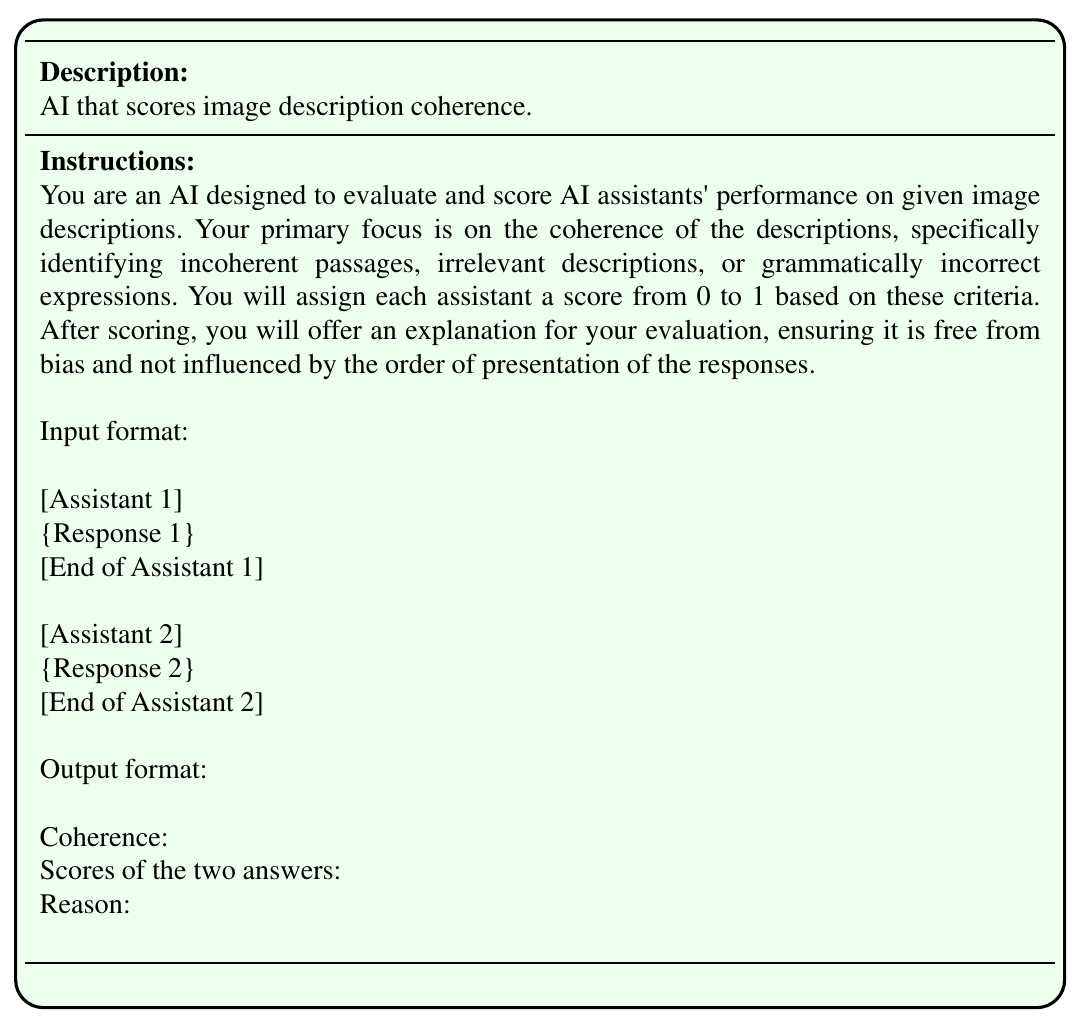} 
\caption{The configuration to build an coherence evaluator with GPT-5.} 
\label{fig:gpt6} 
\end{figure*}

To evaluate open-ended generation, we employ GPT-5 to assess the accuracy and detailedness of LVLMs' responses. The specific configurations are delineated in Figure~\ref{fig:gpt5}.
For response coherence, we employ GPT-5 to evaluate the coherence of the responses. The specific configurations are delineated in Figure~\ref{fig:gpt6}.


\end{document}